\documentclass[acmsmall]{acmart}

\usepackage{amsmath,amsfonts}
\usepackage{algorithmic}
\usepackage{algorithm}
\usepackage{array}
\usepackage{textcomp}
\usepackage{stfloats}
\usepackage{url}
\usepackage{verbatim}
\usepackage{graphicx}

\newcommand{\comm}[1]{\iffalse #1 \fi}
\usepackage{colortbl}
\usepackage{soul}
\usepackage{pifont}
\usepackage{lipsum}
\usepackage{hhline}
\usepackage{booktabs}
\usepackage{subcaption}
\usepackage{multirow}
\usepackage{multicol}
\usepackage{xspace}
\usepackage{balance}

\definecolor{airforceblue}{rgb}{0.36, 0.54, 0.66}

\newcommand{\xmark}{\ding{55}}%

\definecolor{Gray}{gray}{0.9}

\usepackage[edges]{forest}
\definecolor{lightcoral}{rgb}{0.94, 0.5, 0.5}
\definecolor{lightgreen}{rgb}{0.56, 0.93, 0.56}

\definecolor{brightlavender}{rgb}{0.75, 0.58, 0.89}
\definecolor{capri}{rgb}{0.0, 0.75, 1.0}
\definecolor{carminepink}{rgb}{0.92, 0.3, 0.26}
\definecolor{celadon}{rgb}{0.67, 0.88, 0.69}
\definecolor{darkpastelgreen}{rgb}{0.01, 0.75, 0.24}

\definecolor{pastelblue}{rgb}{0.68, 0.78, 0.81}
\definecolor{mintgreen}{rgb}{0.6, 0.98, 0.6}
\definecolor{lavender}{rgb}{0.71, 0.49, 0.86}
\definecolor{peach}{rgb}{1.0, 0.9, 0.71}
\definecolor{coral}{rgb}{1.0, 0.5, 0.31}
\definecolor{mauve}{rgb}{0.88, 0.69, 1.0}
\definecolor{lemonyellow}{rgb}{1.0, 0.96, 0.4}

\definecolor{hidden-draw}{RGB}{205, 44, 36}
\definecolor{hidden-blue}{RGB}{194,232,247}
\definecolor{hidden-orange}{RGB}{243,202,120}
\definecolor{hidden-yellow}{RGB}{242,244,193}
\definecolor{tree-level-1}{RGB}{245,20,85}
\definecolor{tree-level-2}{RGB}{246,86,118}
\definecolor{tree-level-3}{RGB}{248,177,193}
\definecolor{tree-leaf}{RGB}{176,230,198}

\definecolor{Self}{RGB}{255,0,128}
\definecolor{Ensemble}{RGB}{0,127,255}
\definecolor{Iterative}{RGB}{153,51,255}

\definecolor{exemplar1}{RGB}{136,98,148}
\definecolor{exemplar2}{RGB}{148,210,242}
\definecolor{knowledge1}{RGB}{249,219,152}
\definecolor{knowledge2}{RGB}{255,245,220}

\definecolor{lighttealblue}{RGB}{41, 157, 143}
\definecolor{lightplum}{RGB}{233, 196, 106}
\definecolor{harvestgold}{RGB}{216, 118, 89}

\usepackage{nicematrix}
\usepackage{makecell}
\usepackage{xspace,mfirstuc,tabulary}
\usepackage{booktabs}
\usepackage{graphicx}
\usepackage{bbding}
\usepackage{multirow}

\usepackage{pifont}
\newrobustcmd{\B}{\bfseries}
\newcommand*\colourcheck[1]{%
  \expandafter\newcommand\csname #1check\endcsname{\textcolor{#1}{\ding{52}}}%
}
\newcommand*\colourcross[1]{%
  \expandafter\newcommand\csname #1cross\endcsname{\textcolor{#1}{\ding{55}}}%
}
\colourcheck{green}
\colourcross{red}
\newcommand{\eg}{\emph{e.g.}}
\newcommand{\ie}{\emph{i.e.}}

\newcommand{\cmark}{\ding{51}}%

\AtBeginDocument{%
  \providecommand\BibTeX{{%
    \normalfont B\kern-0.5em{\scshape i\kern-0.25em b}\kern-0.8em\TeX}}}

\setcopyright{acmlicensed}
\copyrightyear{2025}
\acmYear{2025}
\acmDOI{XXXXXXX.XXXXXXX}

\acmJournal{Preprint}

\begin{document}

\title{Hallucination of Multimodal Large Language Models: A Survey}


\author{Zechen Bai}
\affiliation{%
  \institution{Show Lab, National University of Singapore}
  \streetaddress{4 Engineering Drive 3}
  \city{Singapore}
  \country{Singapore}}
\email{zechenbai@u.nus.edu}

\author{Pichao Wang}
\affiliation{%
  \institution{Amazon AGI}
  \city{Washington}
  \country{USA}}
\email{pichaowang@gmail.com}

\author{Tianjun Xiao}
\affiliation{%
  \institution{AWS Shanghai AI Lab}
  \city{Shanghai}
  \country{China}}
\email{tianjux@amazon.com}

\author{Tong He}
\affiliation{%
  \institution{AWS Shanghai AI Lab}
  \city{Shanghai}
  \country{China}}
\email{htong@amazon.com}

\author{Zongbo Han}
\affiliation{%
  \institution{Show Lab, National University of Singapore}
  \streetaddress{4 Engineering Drive 3}
  \city{Singapore}
  \country{Singapore}}
\email{hanzb1997@gmail.com}

\author{Zheng Zhang}
\affiliation{%
  \institution{AWS Shanghai AI Lab}
  \city{Shanghai}
  \country{China}}
\email{zhaz@amazon.com}

\author{Mike Zheng Shou}
\affiliation{%
  \institution{Show Lab, National University of Singapore}
  \streetaddress{4 Engineering Drive 3}
  \city{Singapore}
  \country{Singapore}}
  \authornote{Corresponding Author}
\email{mike.zheng.shou@gmail.com}

\renewcommand{\shortauthors}{Bai, et al.}

\begin{abstract}
This survey presents a comprehensive analysis of the phenomenon of hallucination in multimodal large language models (MLLMs), also known as Large Vision-Language Models (LVLMs), which have demonstrated significant advancements and remarkable abilities in multimodal tasks. Despite these promising developments, MLLMs often generate outputs that are inconsistent with the visual content, a challenge known as hallucination, which poses substantial obstacles to their practical deployment and raises concerns regarding their reliability in real-world applications. This problem has attracted increasing attention, prompting efforts to detect and mitigate such inaccuracies. We review recent advances in identifying, evaluating, and mitigating these hallucinations, offering a detailed overview of the underlying causes, evaluation benchmarks, metrics, and strategies developed to address this issue. Additionally, we analyze the current challenges and limitations, formulating open questions that delineate potential pathways for future research. By drawing the granular classification and landscapes of hallucination causes, evaluation benchmarks, and mitigation methods, this survey aims to deepen the understanding of hallucinations in MLLMs and inspire further advancements in the field. Through our thorough and in-depth review, we contribute to the ongoing dialogue on enhancing the robustness and reliability of MLLMs, providing valuable insights and resources for researchers and practitioners alike.
Resources are available at: \textcolor{blue}{\url{https://github.com/showlab/Awesome-MLLM-Hallucination}}.
\end{abstract}




\maketitle

\section{Introduction}

Recently, the emergence of large language models (LLMs) \cite{2022chatgpt, 2023bard, touvron2023llama, penedo2023refinedweb, zhao2023survey} has dominated a wide range of tasks in natural language processing (NLP), achieving unprecedented progress in language understanding \cite{hendrycks2020measuring, huang2023c}, generation \cite{zhang2023benchmarking, zhu2023multilingual} and reasoning \cite{wei2022chain, kojima2022large, qiao2022reasoning, yu2023nature, chu2023survey}.
Leveraging the capabilities of robust LLMs, multimodal large language models (MLLMs)~\cite{llava,minigpt,mplug,instructblip}, sometimes referred to as large vision-language models (LVLMs), are attracting increasing attention.
MLLMs show promising ability in multimodal tasks, such as image captioning~\cite{blip-2}, visual question answering~\cite{instructblip,llava}, etc.
However, there is a concerning trend associated with the rapid advancement in MLLMs. 
These models exhibit an inclination to generate hallucinations~\cite{pope,LURE_zhou2023analyzing,lvlm_survey}, resulting in seemingly plausible yet factually spurious content.

The problem of hallucination originates from LLMs themselves.
In the NLP community, the hallucination problem is empirically categorized into two types~\cite{LLM_survey}:
1) \textit{factuality hallucination} emphasizes the discrepancy between generated content and verifiable real-world facts, typically manifesting as factual inconsistency or fabrication;
2) \textit{faithfulness hallucination} refers to the divergence of generated content from user instructions or the context provided by the input, as well as self-consistency within generated content.
In contrast to pure LLMs, research efforts of hallucination in MLLMs mainly focus on the discrepancy between generated \textbf{text response} and provided \textbf{visual content}~\cite{pope,LURE_zhou2023analyzing,lvlm_survey}, \ie, cross-modal inconsistency.
This difference suggests that studies in LLMs cannot be seemingly transferred to MLLMs.
Therefore, there is a growing need to comprehensively survey recent advancements in MLLMs' hallucination phenomena to inspire new ideas and foster the field's development.

In the realm of computer vision, object recognition is the core task, including sub-tasks such as object classification~\cite{alexnet}, detection~\cite{fastrcnn}, and segmentation~\cite{maskrcnn}, etc.
Similarly, studies on hallucination in MLLMs primarily focus on object hallucination.
In pre-MLLM era, there is a pioneering work on object hallucination in image captioning~\cite{chair}, evaluating object existence by comparing captions and image content.
In MLLMs, object hallucination has been empirically categorized into three categories: 1) \textit{category}, which identifies nonexistent or incorrect object categories in the given image; 2) \textit{attribute}, which emphasizes descriptions of the objects’ attributes, such as color, shape, material, etc; and 3) \textit{relation}, which assesses the relationships among objects, such as human-object interactions or relative positions.
Note that some literature may consider objects counting, objects event, etc., as independent hallucination categories;
however, in this work, we include them into \textit{attribute} category.

As numerous studies exist on the underlying causes of hallucinations in LLMs, the unique challenges posed by cutting-edge MLLMs warrant an in-depth investigation.
Our analysis specifically targets the unique origins of hallucinations in MLLMs, spanning a spectrum of contributing factors from data, model, training, to the inference stage.
In addition, we provide a comprehensive overview of benchmarks and metrics designed specifically for evaluating hallucinations in MLLMs.
Then, we review and discuss recent works tailored to mitigate the problem of hallucination from the viewpoints of the identified causes.

Through our comprehensive survey, we aim to contribute to advancing the field of MLLMs and offer valuable insights that deepen understanding of the opportunities and challenges associated with hallucinations in MLLMs.
This exploration not only enhances our understanding of the limitations of current MLLMs but also offers essential guidance for future research and the development of more robust and trustworthy MLLMs.

\textbf{Comparison with existing surveys.}
In pursuit of reliable generative AI, hallucination stands out as a major challenge, leading to a series of survey papers on its recent advancements.
For pure LLMs, there are several surveys~\cite{zhang2023siren,LLM_survey}, describing the landscape of hallucination in LLMs.
In contrast, there are very few surveys on hallucination in the field of MLLMs.
To the best of our knowledge, there is only one concurrent work~\cite{lvlm_survey}, a short survey on the hallucination problem of LVLMs.
However, our survey distinguishes itself in terms of both taxonomy and scope.
We present a layered and granular classification of hallucinations, as shown in Fig.~\ref{fig:categorization_of_survey}, drawing a clearer landscape of this field.
Additionally, our approach does not limit itself to specific model architectures as prescribed in the work of \cite{lvlm_survey}, but rather dissects the causes of hallucinations by tracing back to various affecting factors.
We cover a larger range of literature both in terms of paper number and taxonomy structure.
Furthermore, our mitigation strategies are intricately linked to the underlying causes, ensuring a cohesive and targeted approach.

\textbf{Organization of this survey.}
In this paper, we present a comprehensive survey of the latest developments regarding hallucinations in MLLMs.
The survey is organized as follows:
We begin by providing sufficient context and defining concepts related to LLMs, MLLMs, hallucination, etc.
Next, we delve into an in-depth analysis of the factors contributing to hallucinations in MLLMs.
Following this, we present a set of metrics and benchmarks employed for evaluating hallucinations in MLLMs.
We then elaborate on a range of approaches designed to mitigate hallucinations in MLLMs.
Finally, we delve into the challenges and open questions that frame the current limitations and future prospects of this field, offering insights and delineating potential pathways for forthcoming research.

\tikzstyle{my-box}=[
    rectangle,
    rounded corners,
    text opacity=1,
    minimum height=1.5em,
    minimum width=5em,
    inner sep=2pt,
    align=center,
    fill opacity=.5,
]
\tikzstyle{cause_leaf}=[my-box, minimum height=1.5em,
    fill=lighttealblue!20, text=black, align=left,font=\scriptsize,
    inner xsep=2pt,
    inner ysep=4pt,
]
\tikzstyle{detect_leaf}=[my-box, minimum height=1.5em,
    fill=lightplum!20, text=black, align=left,font=\scriptsize,
    inner xsep=2pt,
    inner ysep=4pt,
]
\tikzstyle{mitigate_leaf}=[my-box, minimum height=1.5em,
    fill=harvestgold!20, text=black, align=left,font=\scriptsize,
    inner xsep=2pt,
    inner ysep=4pt,
]
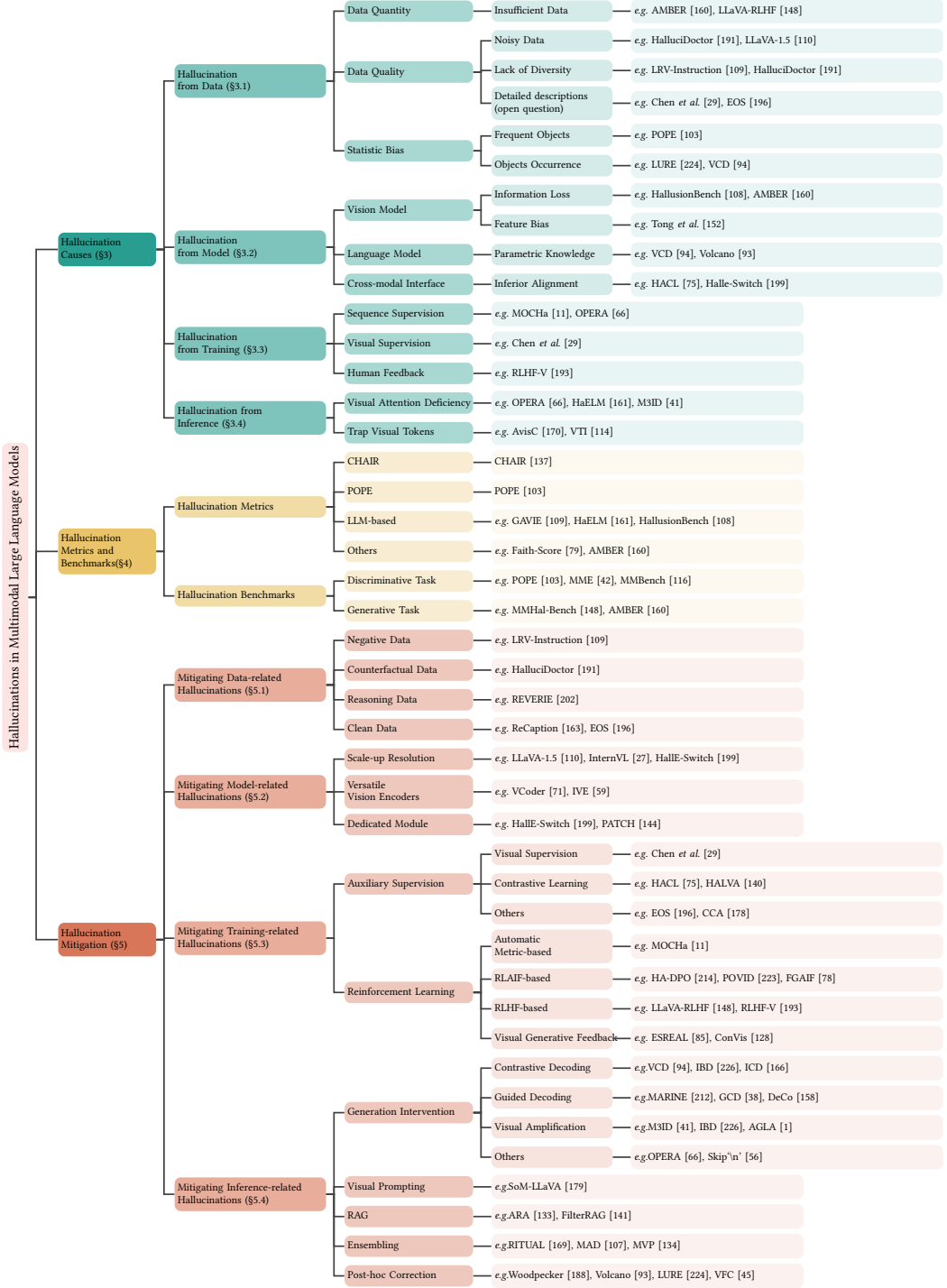
\begin{figure*}[tp]
    \centering
    \resizebox{\textwidth}{!}{
        \begin{forest}
            forked edges,
            for tree={
                grow=east,
                reversed=true,
                anchor=base west,
                parent anchor=east,
                child anchor=west,
                base=left,
                font=\small,
                rectangle,
                rounded corners,
                align=left,
                minimum width=4em,
                edge+={darkgray, line width=1pt},
                s sep=3pt,
                inner xsep=2pt,
                inner ysep=3pt,
                ver/.style={rotate=90, child anchor=north, parent anchor=south, anchor=center},
            },
            where level=1{text width=6.0em,font=\scriptsize,}{},
            where level=2{text width=9.5em,font=\scriptsize,}{},
            where level=3{text width=8.0em,font=\scriptsize,}{},
            where level=4{text width=7.5em,font=\scriptsize,}{},
            [
                Hallucinations in Multimodal Large Language Models, ver, color=carminepink!100, fill=carminepink!15,
                text=black
                [
                    Hallucination \\ Causes (\S \ref{sec:causes}), color=lighttealblue!100, fill=lighttealblue!100, text=black
                    [
                        Hallucination \\ from Data (\S \ref{sec:causes_data}), color=lighttealblue!100, fill=lighttealblue!60,  text=black
                        [
                            Data Quantity, color=lighttealblue!100, fill=lighttealblue!40, text=black
                            [
                                Insufficient Data, color=lighttealblue!100, fill=lighttealblue!20, text=black
                                [
                                    {\eg~AMBER~\cite{amber}, LLaVA-RLHF~\cite{llava_rlhf_sun2023aligning}}
                                	, cause_leaf, text width=20em
                                ]     
                            ] 
                        ]
                        [
                            Data Quality, color=lighttealblue!100, fill=lighttealblue!40, text=black
                            [
                            		Noisy Data, color=lighttealblue!100, fill=lighttealblue!20, text=black
                            		[
                                		{\eg~HalluciDoctor~\cite{yu2023hallucidoctor}, LLaVA-1.5~\cite{llava15}}
                                		, cause_leaf, text width=20em
                            		]
                            ]
                            [
                            		Lack of Diversity, color=lighttealblue!100, fill=lighttealblue!20, text=black
                            		[
                                		{\eg~LRV-Instruction~\cite{LRV_instruction}, HalluciDoctor~\cite{yu2023hallucidoctor}}
                                		, cause_leaf, text width=20em
                            		]
                            ] 
                            [
                            		Detailed descriptions \\ (open question), color=lighttealblue!100, fill=lighttealblue!20, text=black
                            		[
                                		{\eg~Chen~\textit{et al.}~\cite{RAH_bench}, EOS~\cite{eos_token}}
                                		, cause_leaf, text width=20em
                            		]
                            ] 
                        ]
                        [
                            Statistic Bias, color=lighttealblue!100, fill=lighttealblue!40, text=black
                            [
                                Frequent Objects, color=lighttealblue!100, fill=lighttealblue!20, text=black
                                [
                                    {\eg~POPE~\cite{pope}}
                                    , cause_leaf, text width=20em
                                ]
                            ]
                            [
                                Objects Occurrence, color=lighttealblue!100, fill=lighttealblue!20, text=black
                                [
                                    {\eg~LURE~\cite{LURE_zhou2023analyzing}, VCD~\cite{VCD}}
                                    , cause_leaf, text width=20em
                                ]
                            ]  
                        ]
                    ]
                    [
                        Hallucination \\ from Model (\S \ref{sec:causes_model}), color=lighttealblue!100, fill=lighttealblue!60, text=black
                        [
                            Vision Model, color=lighttealblue!100, fill=lighttealblue!40, text=black
                            [
                                Information Loss, color=lighttealblue!100, fill=lighttealblue!20, text=black
                                [
                                    {\eg~HallusionBench~\cite{liu2023hallusionbench}, AMBER~\cite{amber}}
                                    , cause_leaf, text width=20em
                                ]
                            ]
                            [
                                Feature Bias, color=lighttealblue!100, fill=lighttealblue!20, text=black
                                [
                                    {\eg~Tong \textit{et al.}~\cite{tong2024eyes}}
                                    , cause_leaf, text width=20em
                                ]
                            ]
                        ]
                        [
                            Language Model, color=lighttealblue!100, fill=lighttealblue!40, text=black
                            [
                                Parametric Knowledge, color=lighttealblue!100, fill=lighttealblue!20, text=black
                                [
                                    {\eg~VCD~\cite{VCD}, Volcano~\cite{lee2023volcano}}
                                    , cause_leaf, text width=20em
                                ] 
                            ]
                        ]
                        [
                            Cross-modal Interface, color=lighttealblue!100, fill=lighttealblue!40, text=black
                            [
                                Inferior Alignment, color=lighttealblue!100, fill=lighttealblue!20, text=black
                                [
                                    {\eg~HACL~\cite{HACL_contrastive}, Halle-Switch~\cite{halle_switch}}
                                    , cause_leaf, text width=20em
                                ]
                            ]
                        ]
                    ]
                    [
                        Hallucination \\ from Training (\S \ref{sec:causes_training}), color=lighttealblue!100, fill=lighttealblue!60, text=black
                        [
                            Sequence Supervision, color=lighttealblue!100, fill=lighttealblue!40, text=black
                            [
                                {\eg~MOCHa~\cite{ben2023mocha_openchair}, OPERA~\cite{huang2023opera}}
                                , cause_leaf, text width=20em
                            ]
                        ]
                        [
                            Visual Supervision, color=lighttealblue!100, fill=lighttealblue!40, text=black
                            [
                                {\eg~Chen~\textit{et al.}~\cite{RAH_bench}}
                                , cause_leaf, text width=20em
                            ]
                        ]
                        [
                            Human Feedback, color=lighttealblue!100, fill=lighttealblue!40, text=black
                            [
                                {\eg~RLHF-V~\cite{yu2023rlhf_v}}
                                , cause_leaf, text width=20em
                            ]
                        ]
                    ]
                    [
                        Hallucination from \\ Inference (\S \ref{sec:causes_infer}), color=lighttealblue!100, fill=lighttealblue!60, text=black
                        [
                            Visual Attention Deficiency, color=lighttealblue!100, fill=lighttealblue!40, text=black
                            [
                                {\eg~OPERA~\cite{huang2023opera}, HaELM~\cite{haelm}, M3ID~\cite{M3ID}}
                                , cause_leaf, text width=20em
                            ]
                        ]
                        [
                            Trap Visual Tokens, color=lighttealblue!100, fill=lighttealblue!40, text=black
                            [
                                {\eg~AvisC~\cite{avisc}, VTI~\cite{vti_iclr25}}
                                , cause_leaf, text width=20em
                            ]
                        ]
                    ] 
                ]
                [
                    Hallucination \\ Metrics and \\ Benchmarks(\S \ref{sec:detection_and_benchmark}), color=lightplum!100, fill=lightplum!100, text=black
                    [
                        Hallucination Metrics, color=lightplum!100, fill=lightplum!60, text=black
                        [
                            CHAIR, color=lightplum!100, fill=lightplum!30, text=black
                            [
                                {CHAIR~\cite{chair}}
                                , detect_leaf, text width=20em
                            ]
                        ]
                        [
                            POPE, color=lightplum!100, fill=lightplum!30, text=black
                            [
                                {POPE~\cite{pope}}
                                , detect_leaf, text width=20em
                            ]
                        ]
                        [
                            LLM-based, color=lightplum!100, fill=lightplum!30, text=black
                            [
                                {\eg~GAVIE~\cite{LRV_instruction}, HaELM~\cite{haelm}, HallusionBench~\cite{liu2023hallusionbench}}
                                , detect_leaf, text width=20em
                            ]
                        ]
                        [
                            Others, color=lightplum!100, fill=lightplum!30, text=black
                            [
                                {\eg~Faith-Score~\cite{jing2023faithscore}, AMBER~\cite{amber}}
                                    , detect_leaf, text width=20em
                            ]
                        ]
                    ]
                    [
                        Hallucination Benchmarks, color=lightplum!100, fill=lightplum!60, text=black
                        [
                            Discriminative Task, color=lightplum!100, fill=lightplum!30, text=black
                            [
                                {\eg~POPE~\cite{pope}, MME~\cite{fu2023mme}, MMBench~\cite{liu2024mmbench}}
                                , detect_leaf, text width=20em
                            ]
                        ]
                        [
                            Generative Task, color=lightplum!100, fill=lightplum!30, text=black
                            [
                                {\eg~MMHal-Bench~\cite{llava_rlhf_sun2023aligning}, AMBER~\cite{amber}}
                                , detect_leaf, text width=20em
                            ]
                        ]
                    ]
                ]
                [
                    Hallucination \\ Mitigation (\S \ref{sec:mitigating}), color=harvestgold!100, fill=harvestgold!100, text=black
                    [
                        Mitigating Data-related \\ Hallucinations (\S \ref{sec:mitigating_data}), color=harvestgold!100, fill=harvestgold!60, text=black
                        [
                            Negative Data, color=harvestgold!100, fill=harvestgold!40, text=black
                            [
                                {\eg~LRV-Instruction~\cite{LRV_instruction}},mitigate_leaf, text width=20em
                            ]
                        ]
                        [
                            Counterfactual Data, color=harvestgold!100, fill=harvestgold!40, text=black
                            [
                                {\eg~HalluciDoctor~\cite{yu2023hallucidoctor}},mitigate_leaf, text width=20em
                            ]
                        ]
                        [
                            Reasoning Data, color=harvestgold!100, fill=harvestgold!40, text=black
                            [
                                {\eg~REVERIE~\cite{zhang2024reflective}},mitigate_leaf, text width=20em
                            ]
                        ]
                        [
                            Clean Data, color=harvestgold!100, fill=harvestgold!40, text=black
                            [
                                {\eg~ReCaption~\cite{fghe_wang2023mitigating}, EOS~\cite{eos_token}},mitigate_leaf, text width=20em
                            ]
                        ]
                    ]
                    [
                        Mitigating Model-related \\ Hallucinations (\S \ref{sec:mitigating_model}), color=harvestgold!100, fill=harvestgold!60, text=black
                        [
                            Scale-up Resolution, color=harvestgold!100, fill=harvestgold!40, text=black
                            [
                                {\eg~LLaVA-1.5~\cite{llava15}, InternVL~\cite{chen2024internvl}, HallE-Switch~\cite{halle_switch}}, mitigate_leaf, text width=20em
                            ]
                        ]
                        [
                            Versatile \\ Vision Encoders, color=harvestgold!100, fill=harvestgold!40, text=black
                            [
                                {\eg~VCoder~\cite{jain2023vcoder}, IVE~\cite{visual_experts}}, mitigate_leaf, text width=20em
                            ]
                        ]
                        [
                            Dedicated Module, color=harvestgold!100, fill=harvestgold!40, text=black
                            [
                                {\eg~HallE-Switch~\cite{halle_switch}, PATCH~\cite{PATCH}}, mitigate_leaf, text width=20em
                            ]
                        ]
                    ]
                    [
                        Mitigating Training-related \\ Hallucinations (\S \ref{sec:mitigating_training}), color=harvestgold!100, fill=harvestgold!60, text=black
                        [
                            Auxiliary Supervision, color=harvestgold!100, fill=harvestgold!40, text=black
                            [
                                Visual Supervision, color=harvestgold!100, fill=harvestgold!20, text=black
                                [
                                    {\eg~Chen \textit{et al.}~\cite{RAH_bench}}, mitigate_leaf, text width=20em
                                ]
                            ]
                            [
                                Contrastive Learning, color=harvestgold!100, fill=harvestgold!20, text=black
                                [
                                    {\eg~HACL~\cite{HACL_contrastive}, HALVA~\cite{halva}},mitigate_leaf, text width=20em
                                ]
                            ]
                            [
                                Others, color=harvestgold!100, fill=harvestgold!20, text=black
                                [
                                    {\eg~EOS~\cite{eos_token}, CCA~\cite{CCA}},mitigate_leaf, text width=20em
                                ]
                            ]
                        ]
                        [
                            Reinforcement Learning, color=harvestgold!100, fill=harvestgold!40, text=black
                            [
                                {Automatic\\ Metric-based}, color=harvestgold!100, fill=harvestgold!20, text=black
                                [
                                    {\eg~MOCHa~\cite{ben2023mocha_openchair}},mitigate_leaf, text width=20em
                                ]
                            ]
                            [
                                RLAIF-based, color=harvestgold!100, fill=harvestgold!20, text=black
                                [
                                    {\eg~HA-DPO~\cite{HA_DPO}, POVID~\cite{dpo_povid}, FGAIF~\cite{jing2024fgaif}},mitigate_leaf, text width=20em
                                ]
                            ]
                            [
                                RLHF-based, color=harvestgold!100, fill=harvestgold!20, text=black
                                [
                                    {\eg~LLaVA-RLHF~\cite{llava_rlhf_sun2023aligning}, RLHF-V~\cite{yu2023rlhf_v}},mitigate_leaf, text width=20em
                                ]
                            ]
                            [
                                Visual Generative Feedback, color=harvestgold!100, fill=harvestgold!20, text=black
                                [
                                    {\eg~ESREAL~\cite{ESREAL}, ConVis~\cite{park2024convis}},mitigate_leaf, text width=20em
                                ]
                            ]
                        ]
                    ]
                    [
                        Mitigating Inference-related \\ Hallucinations (\S \ref{sec:mitigating_inference}), color=harvestgold!100, fill=harvestgold!60, text=black
                        [
                            Generation Intervention, color=harvestgold!100, fill=harvestgold!40, text=black
                            [
                                Contrastive Decoding, color=harvestgold!100, fill=harvestgold!20, text=black
                                [
                                    {\eg VCD~\cite{VCD}, IBD~\cite{zhu2024ibd}, ICD~\cite{ICD}},mitigate_leaf, text width=20em
                                ]
                            ]
                            [
                                Guided Decoding, color=harvestgold!100, fill=harvestgold!20, text=black
                                [
                                    {\eg MARINE~\cite{cfg_mitigate}, GCD~\cite{deng2024seeing}, DeCo~\cite{deco}}, mitigate_leaf, text width=20em
                                ]
                            ]
                            [
                                Visual Amplification, color=harvestgold!100, fill=harvestgold!20, text=black
                                [
                                    {\eg M3ID~\cite{M3ID}, IBD~\cite{zhu2024ibd}, AGLA~\cite{an2024agla}}, mitigate_leaf, text width=20em
                                ]
                            ]
                            [
                                Others, color=harvestgold!100, fill=harvestgold!20, text=black
                                [
                                    {\eg OPERA~\cite{huang2023opera}, Skip`\textbackslash{}n'~\cite{han2024skip}}, mitigate_leaf, text width=20em
                                ]
                            ]
                        ]
                        [
                            Visual Prompting, color=harvestgold!100, fill=harvestgold!40, text=black
                            [
                                {\eg SoM-LLaVA~\cite{som_llava}}
                                , mitigate_leaf, text width=20em
                            ]
                        ]
                        [
                            RAG, color=harvestgold!100, fill=harvestgold!40, text=black
                            [
                                {\eg ARA~\cite{ara}, FilterRAG~\cite{filterRAG}}
                                , mitigate_leaf, text width=20em
                            ]
                        ]
                        [
                            Ensembling, color=harvestgold!100, fill=harvestgold!40, text=black
                            [
                                {\eg RITUAL~\cite{woo2024ritual}, MAD~\cite{multi_agent_debate}, MVP~\cite{MVP}}
                                , mitigate_leaf, text width=20em
                            ]
                        ]
                        [
                            Post-hoc Correction, color=harvestgold!100, fill=harvestgold!40, text=black
                            [
                                {\eg Woodpecker~\cite{yin2023woodpecker}, Volcano~\cite{lee2023volcano}, LURE~\cite{LURE_zhou2023analyzing}, VFC~\cite{VFC}}
                                , mitigate_leaf, text width=20em
                            ]
                        ]
                    ]
                ]
            ]
        \end{forest}
    }
    \caption{The main content flow and categorization of this survey.}
    \label{fig:categorization_of_survey}
\end{figure*}

\section{Definitions}
\label{sec:definition}

\subsection{Large Language Models}
Before moving to multimodal large language models, it is essential to introduce the concept of large language models.
Typically, LLMs encompass a range of transformer-based models that are extensively trained on vast textual datasets.
Prominent examples include GPT-3~\cite{brown2020language}, PaLM~\cite{chowdhery2022palm}, LLaMA~\cite{touvron2023llama}, and GPT-4~\cite{GPT4_openai}.
Through scaling both data volume and model capacity, LLMs demonstrate notable emergent capabilities, including In-Context Learning\cite{brown2020language}, Chain-of-Thought prompting\cite{wei2022chain} and instruction following\cite{peng2023instruction}, among others.

The characteristics and behaviors of LLMs are intricately linked to their training processes. LLMs typically undergo three primary training stages: pre-training, Supervised Fine-Tuning (SFT), and Reinforcement Learning from Human Feedback (RLHF). Below, we provide a concise overview of each stage to facilitate comprehension.

\textbf{Pre-trianing.}
Pre-training serves as a fundamental phase in the learning process of LLMs~\cite{zhou2023lima}.
During this stage, language models engage in autoregressive prediction, wherein they predict the subsequent token in a sequence.
By undergoing self-supervised training on vast textual datasets, these models develop an understanding of language syntax, gain access to world knowledge, and enhance their reasoning capabilities.
This pre-training process establishes a solid groundwork for the models to undertake subsequent fine-tuning tasks effectively.

\textbf{Supervised Fine-Tuning.}
Although pre-training equips LLMs with substantial knowledge and skills, it's important to acknowledge that its primary focus is on optimizing for completion.
Consequently, pre-trained LLMs essentially function as completion machines, which may create a misalignment between the objective of predicting the next word within LLMs and the user's objective of obtaining desired responses.
To address this disparity, the concept of Supervised Fine-Tuning (SFT)~\cite{zhang2023instruction} has been introduced.
SFT involves further training LLMs using a meticulously annotated set of (instruction, response) pairs, thereby enhancing the capabilities and controllability of LLMs.

\textbf{Reinforcement Learning from Human Feedback.}
Although SFT has made strides in enabling LLMs to adhere to user instructions, there remains a need for further alignment with human preferences.
Among the various methods, Reinforcement Learning from Human Feedback (RLHF)~\cite{christiano2017deep, stiennon2020learning,ouyang2022training} emerges as a notable approach for achieving alignment through reinforcement learning.
RLHF typically employs a preference model~\cite{bradley1952rank}, trained to predict preference rankings based on prompts and human-labeled responses.
To better align with human preferences, RLHF optimizes the LLM to generate outputs that maximize rewards provided by the trained preference model, often utilizing reinforcement learning algorithms like Proximal Policy Optimization (PPO)~\cite{ppo}.
This integration of human feedback into the training loop has demonstrated effectiveness in enhancing the alignment of LLMs.

\begin{figure}[t]
    \centering
    \includegraphics[width=0.65\linewidth]{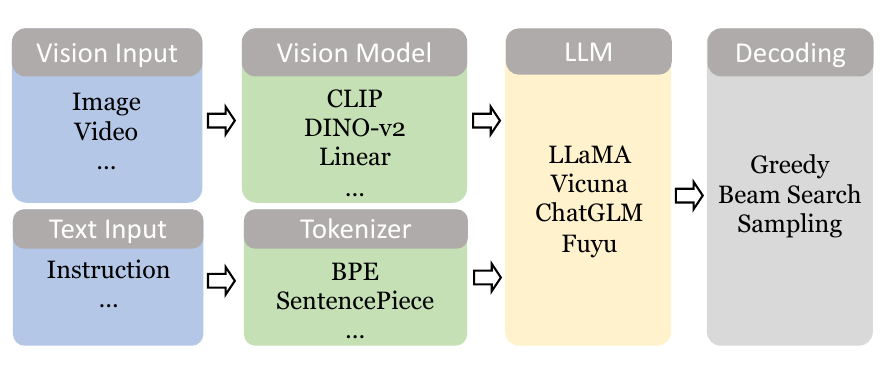}
    \caption{Popular architecture of multimodal large language model.}
    \label{fig:mllm_arch}
\end{figure}

\subsection{Multimodal Large Language Models}
\label{sec:mllm_training}
MLLMs \cite{llava,minigpt,mplug,instructblip,video_lisa,zhao2024lova3} typically refers to a series of models that enable LLMs to perceive and comprehend data from various modalities.
Among them, vision+LLM is particularly prominent, owing to the extensive research on vision-language models (VLMs) \cite{CLIP, ALIGN, yu2205coca} prior to LLMs.
As a result, MLLMs are sometimes referred to as vision-LLMs (VLLMs) or large vision language models (LVLMs).
The goal of MLLMs is to activate the visual capabilities of LLMs, enabling them to "see" the world via images or videos.
Combined with strong reasoning and language generation abilities, MLLMs trigger a series of downstream tasks in multimodal domains, such as image/video captioning~\cite{wang2020show,bai2021explain,fan2023unsupervised}, visual question answering, cross-modal retrieval~\cite{bai2025bridging}, and so on.
Additionally, MLLMs serve as the foundation for applications in other fields, such as AI assistants, embodied agents, and robotics.
Recently, there is a trend of developing unified multimodal models~\cite{showo,zhou2024transfusion,chen2025janus} that integrates both understanding and generation, which are essentially visual-to-text and text-to-visual generation~\cite{stable_diffusion,fqgan,bai2025impossible}.
In this survey, we focus on the multimodal understanding task, i.e., visual-to-text generation.

Integrating the two modalities of vision and language involves primarily two types of approaches.
The first line of work is built upon off-the-shelf pre-trained uni-modal models.
Specifically, these MLLMs usually incorporate a learnable interface between pre-trained visual encoders and LLMs.
The interface extracts and integrates information from visual modalities.
Such interfaces can be further categorized into 1) learnable query-based and 2) projection layer based.
Learnable query-based methods, exemplified by Q-Former~\cite{blip-2}, as used in MiniGPT-4~\cite{minigpt} and Instruct-BLIP~\cite{instructblip}, utilize a set of learnable query tokens to capture visual signals via cross-attention.
Projection layer-based methods, as widely applied in LLaVA~\cite{llava}, Shikra~\cite{chen2023shikra}, etc., involve training a linear projection layer or a Multi-Layer Perceptron (MLP) module to transform extracted visual features.
Both types of interfaces aim to transform pre-trained visual features into the input space of pre-trained LLMs.

Another line of work is represented by Fuyu-8B~\cite{fuyu8b} and Gemini~\cite{team2023gemini}.
Unlike previous methods that leverage pre-trained uni-modal models, these works employ end-to-end training from scratch.
Taking Fuyu-8B as an example, it does not employ any pre-trained vision encoder.
Instead, it directly inputs image patches and employs a linear projection to transform the raw pixels of each patch into embeddings.

The abstracted pipeline is depicted in Fig.~\ref{fig:mllm_arch}.
MLLMs take input from both visual and textual modalities, learning from multimodal instructions and responses, which leads to remarkable performance across various multimodal tasks.
Regarding the training of MLLMs, we provide a concise overview of the training process for interface-based MLLMs.
Given that end-to-end models are closed-source, the training details are unknown.
Typically, the training of interface-based MLLMs consists of two stages: 1) pre-training, 2) instruction tuning.

\textbf{Pre-training.}
Given that models from each modality are pre-trained on their respective data, the objective of this pre-training phase is to achieve cross-modal feature alignment.
During training, both the pre-trained visual encoder and LLM remain frozen, with only the cross-modal interface being trained.
Similar to traditional VLMs training, as exemplified by CLIP~\cite{CLIP}, web-scale image-text pairs \cite{schuhmann2022laion} are utilized for training.
Given that the final output is at the LLM side, the most widely used loss function in this stage is the text generation loss, typically cross-entropy loss, which aligns with the pre-training of LLMs.
Certain studies (e.g., \cite{blip-2,instructblip}) explore the incorporation of contrastive loss and image-text matching loss to further enhance alignment.
After training, the interface module maps the visual features into the input embedding space of the LLM.

\textbf{Instruction Tuning.}
Similar to LLMs, after pre-training, the current model still lacks instruction following ability in the multimodal context.
During the instruction tuning stage, both machine-generated datasets~\cite{llava} and human-annotated QA datasets~\cite{hudson2019gqa,mishra2019ocr,visual_genome} are utilized to enhance the model's ability to comprehend and follow multimodal instructions.
Unlike pre-training data, the format and quality of instruction tuning data significantly impact the model's performance.
It is usually in the format of \textit{visual content - instruction - response}.
Empirical studies also demonstrate that high-quality data significantly enhances the performance of MLLMs.
During this stage, there are various options for training, such as fine-tuning LLM parameters in full~\cite{llava}, or using techniques like LoRA~\cite{hu2021lora} to tune specific LLM parameters.

\subsection{Hallucinations in Multimodal Large Language Models}
\begin{figure*}[t]
    \centering
    \includegraphics[width=0.95\linewidth]{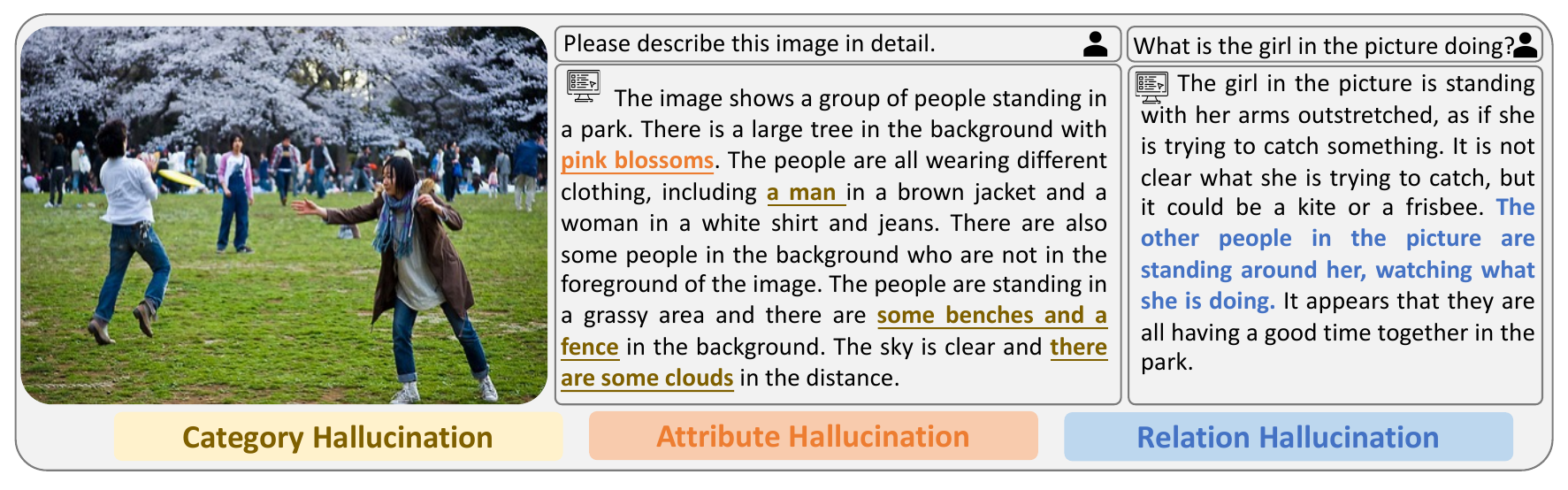}
    \caption{Three types of typical hallucination.}
    \label{fig:hallucination_type}
\end{figure*}

Hallucination of MLLM generally refers to the phenomenon where the generated text response does not align with the corresponding visual content.
State-of-the-art studies in this field primarily focus on object hallucination, given that objects are central to research in computer vision and multimodal contexts.
Regarding inconsistency, two typical failure modes are: 1) missing objects, and 2) describing objects that are not present in the image or with incorrect statements.
Empirically, the second mode has been shown to be less preferable to humans.
For example, the LSMDC challenge~\cite{lsmdc_challenge} shows that correctness is more important to human judges than specificity.
In contrast, the coverage of objects is less perceptible to humans.
Thus, object coverage is not a primary focus in studies of object hallucination.
Empirically, object hallucination can be categorized into three types: \textit{object category}, \textit{object attribute}, and \textit{object relation}.
An example of the three types of hallucination is shown in Fig.~\ref{fig:hallucination_type}.

\begin{itemize}
    \item Category.
    MLLMs identify nonexistent object categories or incorrect categories in the given image.
    For example, in Fig.~\ref{fig:hallucination_type}, "some benches and a fence", "some clouds", described in the text response do not exist in the given image.
    
    \item Attribute.
    The object categories identified by MLLMs are accurate, while the descriptions of these objects’ attributes (such as color, shape, material, content, counting, action, etc.) are wrong.
    In Fig.~\ref{fig:hallucination_type}, "pink blossoms" is hallucinated by the MLLM as the color is inaccurate.
    
    \item Relation. All objects and their attributes are described correctly, but the relationships among them (such as human-object interactions or relative positions) do not align with the actual image content.
    In Fig.~\ref{fig:hallucination_type}, "...standing around her, watching..." is a typical example of relation hallucination, as the objects are presented in the image but the relation is inaccurate.
\end{itemize}

It's worth noting that some literature may categorize objects counting, objects event, etc., as independent hallucination categories.
In this work, we classify them under the \textit{attribute} category.
Some works also discuss `verb' hallucination~\cite{wang2024verb}, which might falls under the domain of either attribute or relation.
Thus, we do not discuss it independently.
The definition of hallucination types aligns well with the domain of compositional generalization~\cite{aro,ma2023crepe} of VLMs, which investigates visio-linguistic generalization and reasoning abilities.

\section{Hallucination Causes}
\label{sec:causes}

Hallucinations have multifaceted origins, spanning the entire spectrum of MLLMs' capability acquisition process.
In this section, we delve into the root causes of hallucinations in MLLMs, primarily categorized into four aspects: \textit{Data}, \textit{Model}, \textit{Training}, and \textit{Inference}.

\subsection{Data}
\label{sec:causes_data}
Data stands as the bedrock for MLLMs, enabling them to gain cross-modal understanding and instruction-following capabilities.
However, it can inadvertently become the source of MLLM hallucinations. 
This mainly manifests in three aspects: quantity, quality, and statistical bias.

\subsubsection{Quantity}
Deep learning models are data-hungry, especially large models like MLLMs.
The amount of data plays an important role in building robust and reliable MLLMs.
Currently, image-text pair datasets~\cite{schuhmann2022laion} and visual QA~\cite{hudson2019gqa,mishra2019ocr} data are used for training MLLMs.
Although these datasets are usually larger than typical datasets in computer vision, they are still far less abundant than the text-only data used for training LLMs in terms of quantity.
Insufficient data could potentially lead to problematic cross-modal alignment, resulting in hallucinations~\cite{amber,llava_rlhf_sun2023aligning}.

\subsubsection{Quality}
Given the increasing demand for large-scale training data, heuristic data collection methods are employed to efficiently gather vast volumes of data.
While these methods provide extensive data, they offer no guarantee of quality, thereby increasing the risk of hallucinations.
Data quality relevant to hallucinations can be further categorized into the following three facets.

\begin{itemize}
    \item
    \textbf{Noisy data.}
    As mentioned in the definition section, training MLLMs involves two stages.
    The pre-training stage employs image-text pairs crawled from the web, which contain inaccurate, misaligned, or corrupted data samples.
    The noisy data would limit the cross-modal feature alignment~\cite{yu2023hallucidoctor,eos_token}, which serves as the foundation of MLLMs.
    As for the instruction tuning data, prevalent methods, such as LLaVA~\cite{llava}, utilize the advanced GPT-4~\cite{GPT4_openai} model to generate instructions.
    However, ChatGPT is a language model that cannot interpret visual content, leading to the risk of noisy data.
    Moreover, language models themselves suffer from the issue of hallucination~\cite{LLM_survey}, further increasing the risk.
    LLaVA-1.5~\cite{llava15} adds human annotated QA data into instruction following and shows improved results, revealing the effect of noisy data.

    \item 
    \textbf{Lack of diversity.}
    Recent works~\cite{LRV_instruction,yu2023hallucidoctor} reveal that the diversity of data also plays a crucial role.
    For the data used in the two training stages, instruction tuning data are more likely to have this issue since it is usually in a relatively small amount.
    One prominent property is that most instruction following data samples are composed of conversations regarding the image content.
    We regard this type of data as \textit{positive instruction}, as it always faithfully reflects the image content.
    In contrast, \textit{negative instruction} data~\cite{LRV_instruction} and \textit{reject answering} responses~\cite{vqav2_idk} are rare in the datasets.
    Given such training data, one potential drawback observed by recent studies~\cite{pope,LRV_instruction} is that current models tend to answer "\textit{Yes}" for any instructions presented to the model, even when a proper answer should be "\textit{No}", leading to hallucination.
    This phenomenon indicates the effect of data diversity.

    \item
    \textbf{Detailed descriptions (open question)}
    The impact of the level of detail in textual descriptions on this matter remains an open question.
    As discussed in Sec.~\ref{sec:mllm_training}, the texts in pre-training data, such as LAION~\cite{schuhmann2022laion}, usually describe the salient objects' overall content.
    While the texts in the instructing tuning stage, such as \textsc{LLaVA-150k}~\cite{llava}, consist of more detailed descriptions.
    This \textsc{LLaVA-150k} dataset is generated by GPT-4 based on objects recognized by vision models.
    One recent work~\cite{RAH_bench} argues that within the training data, detailed descriptions related to object position, attributes, and non-salient objects are usually absent.
    This property results in incomplete cross-modal alignment and deprives the model of grounding ability~\cite{zhang2023gpt4roi,lai2023lisa}.
    However, another work~\cite{eos_token} hypothesizes that the text descriptions in the instruction tuning data contain too much details, exceeding the perception limit of MLLMs.
    When trained with such detailed data, in an attempt to fit the detail level and length distribution of ground truth captions, the model may risk expressing details that it cannot discern from the image, and therefore exhibit hallucinations.
    The detail level of the training data remains an open question.
    
\end{itemize}

\subsubsection{Statistic bias.}
Neural networks, especially large language models, possess an intrinsic tendency to memorize training data, as noted in~\cite{compression}.
The nous (\textit{e.g.}, objects) distribution in the training dataset has strong effects on the behavior of the model.
Frequently appeared objects and object co-occurrence are two prominent types of statistical bias, as discussed in~\cite{chair,pope,LURE_zhou2023analyzing}.
For example, `\textit{person}' might be one of the most frequently appearing objects in the training data.
During inference, even if the given image does not contain a person, the model still tends to predict the presence of a person.
On the other hand, object co-occurrence refers to the phenomenon that the model will remember which two objects usually `go together'~\cite{chair}.
For instance, given an image of a kitchen with a refrigerator, MLLMs are prone to answer `\textit{Yes}' when asked about a microwave, as refrigerators and microwaves frequently appear together in kitchen scenes.
Bias exists in most datasets.
Increasing the scale of data may alleviate the effect, but cannot fully resolve it, given the long-tail distribution of the real world.

\subsection{Model}
\label{sec:causes_model}
Currently, the architecture of popular MLLMs is composed of several components, usually including pre-trained vision model, pre-trained LLM, and alignment module as we discussed above.
Since these models are connected together, instead of end-to-end training from scratch, the error of each module can be accumulated.
Inferior and problematic output from each module may lead to hallucinations.

\begin{itemize}
    \item
    \textbf{Weak vision model.}
    As mentioned in related works~\cite{chair,guan2023hallusionbench_mllm,amber}, a primary potential reason for hallucination is a weak vision model, which can lead to misclassification or misinterpretation of visual concepts.
    Even the most powerful vision model may still experience information loss during the encoding process.
    Weak vision model implies weak perception, which fundamentally undermines the multimodal understanding.

    \item
    \textbf{Language model prior.}
    The modern architecture of MLLMs is imbalanced.
    Usually, the language model is much larger and stronger than the vision model, leading to a tendency to prioritize language-based information~\cite{guan2023hallusionbench_mllm,lee2023volcano,VCD,LRV_instruction,chair}.
    A typical phenomenon is that the knowledge entailed in the language model, also termed as parametric knowledge, can override the visual content.
    For example, given an image showing a red banana, which is counter-intuitive in the real world, an MLLM may still respond with "yellow banana", as "banana is yellow" is a deep-rooted knowledge in the LLM.
    Such language/knowledge prior makes the model overlook the visual content and response with hallucination.

    \item 
    \textbf{Weak alignment interface.}
    The alignment interface plays an essential role in MLLMs, as it serves as the bridge between the two modalities.
    A weak alignment interface can easily cause hallucinations.
    One potential cause of a weak alignment interface is data, as discussed in earlier sections.
    Apart from that, the interface architecture itself and training loss design also matter~\cite{HACL_contrastive,halle_switch,NOPE}.
    Recent work~\cite{HACL_contrastive} argues that the LLaVA-like linear projection interface preserves most of the information, but lacks supervision on the projected feature.
    Visualization in~\cite{HACL_contrastive} reveals that the features after the projection layer remain distinct from the language embeddings.
    The distribution gap causes trouble in cross-modal interaction, leading to hallucination.
    On the other hand, Q-former-like~\cite{blip-2} architecture has diverse supervision on the extracted visual feature, aligning it to the language embedding space.
    However, the use of learnable queries inevitably results in the loss of fine-grained visual information.
    
\end{itemize}

\subsection{Training}
\label{sec:causes_training}
The training objective of MLLMs is basically the same as LLMs, \textit{i.e,} auto-regressive next token prediction loss.
This loss is straightforward yet effective and easy to scale up, showing promising performance in language modeling.
However, some studies in the field of MLLMs have suggested that the next-token prediction loss might not be suitable for learning visual content due to its complex spatial structure~\cite{ben2023mocha_openchair,RAH_bench}.
Additionally, the loss optimizes at the token level, while lacking supervision at the sequence level~\cite{ben2023mocha_openchair}.
Another perspective is that, unlike training LLMs, the RLHF stage is absent in training procedure of MLLMs~\cite{yu2023rlhf_v,llava_rlhf_sun2023aligning}, becoming a potential cause of hallucination.

\subsection{Inference}
\label{sec:causes_infer}

\subsubsection{Visual Attention Deficiency}
As for inference, some works also argues a potential issue in the auto-regressive generation.
During generation, as the sequence length grows, the self-attention will focus more on the previously generated text tokens, \textit{i.e.,} the attention on the visual content is diluted~\cite{amber,haelm,huang2023opera,wang2023vigc,M3ID}.
Through visualizing the attention map during generation~\cite{huang2023opera,haelm}, it can be observed that the generated content focuses more on previous special tokens, such as punctuation, rather than visual content tokens.
The issue of 'losing attention' would also lead to the model's output response being irrelevant to the visual content.

\subsubsection{Trap Visual Tokens}
Even within the group of visual tokens, each token may play a different role.
The work of AvisC~\cite{avisc} analyzes that there are a few image tokens receiving excessive attention, while tokens receiving lower attention weights often hold essential information for identifying nuanced object details.
This imbalanced attention distribution among visual tokens can lead to hallucinatory responses in tasks requiring fine-grained understanding of visual objects.
More recently, the work of VTI~\cite{vti_iclr25} finds that vision encoder may generate some tokens that are very sensitive to image noise.
Such noise can produce outliers in the representation space, resulting in large variability in some of the features, hence hallucinations.

Similar findings have also been discussed in some recent work~\cite{che2025eazy,hjijackers}, \textit{i.e.}, a small portion of visual tokens can severely affect the final results and induce hallucination.

\section{Hallucination Metrics and Benchmarks}
\label{sec:detection_and_benchmark}

\begin{table*}[ht!]
    \centering
    \renewcommand*{\arraystretch}{1}
    \caption{Summary of most relevant benchmarks and metrics of object hallucination in MLLMs. The order is based on chronological order on arxiv. In the metric column, Acc/P/R/F1 denotes Accuracy/Precision/Recall/F1-Score.}
    \setlength\tabcolsep{4pt}
    \resizebox{\linewidth}{!}{
    \begin{tabular}{@{}lclccccccc@{}}
        \toprule
        \multirow{2}{*}{Benchmark} & \multirow{2}{*}{Venue} & \multirow{2}{*}{\shortstack[l]{Underlying\\ Data Source}} & \multirow{2}{*}{Size} & \multirow{2}{*}{\shortstack[l]{Task\\ Type}} & \multirow{2}{*}{Metric} & \multicolumn{4}{c}{Hallucination Type} \\ 
        \arrayrulecolor{gray} \cmidrule(lr){7-10} \arrayrulecolor{black}
        & & & & & & Category & Attribute & Relation & Others \\ 
        \midrule

        CHAIR~\cite{chair} & EMNLP'18 & MSCOCO~\cite{mscoco} & 5,000 & Gen & CHAIR & \cmark & \xmark & \xmark & \xmark \\
        POPE~\cite{pope} & EMNLP'23 & MSCOCO~\cite{mscoco} & 3,000 & Dis & Acc/P/R/F1 & \cmark & \xmark & \xmark & \xmark \\
        MME~\cite{fu2023mme} & arXiv'23 Jun & MSCOCO~\cite{mscoco} & 1457 & Dis & Acc/Score & \cmark & \cmark & \xmark & \cmark \\
        MMBench~\cite{liu2024mmbench} & ECCV'24 & Not Specified & 3217 & Dis & Acc & \cmark & \cmark & \cmark & Reasoning \\
        CIEM~\cite{hu2023ciem} & NeurIPS-W'23 & MSCOCO~\cite{mscoco} & 78120 & Dis & Acc & \cmark & \xmark & \xmark & \xmark \\
        M-HalDetect~\cite{HalDectect_gunjal2023detecting} & AAAI'24 & MSCOCO~\cite{mscoco} & 4,000 & Dis & Reward Model Score & \cmark & \xmark & \xmark & \xmark \\
        MMHal-Bench~\cite{llava_rlhf_sun2023aligning} & arXiv'23 Sep. & Open-Images~\cite{openimgs} & 96 & Gen & LLM Assessment & \cmark & \xmark & \xmark & \cmark \\
        GAVIE~\cite{LRV_instruction} & ICLR'24 & Visual-Genome~\cite{visual_genome} & 1,000 & Gen& LLM Assessment & \multicolumn{4}{c}{Not Explicitly Stated} \\
        NOPE~\cite{NOPE} & ACL-W'24 & Open-Images~\cite{openimgs} & 36,000 & Dis & Acc/METEOR~\cite{banerjee2005meteor} & \cmark & \xmark & \xmark & \xmark \\
        HaELM~\cite{haelm} & arXiv'23 Oct. & MSCOCO~\cite{mscoco} & 5,000 & Gen & LLM Assessment & \multicolumn{4}{c}{Not Explicitly Stated} \\
        FaithScore~\cite{jing2023faithscore} & EMNLP'24 & MSCOCO~\cite{mscoco} & 2,000 & Gen & FaithScore & \cmark & \cmark & \cmark & Obj. Counting \\
        Bingo~\cite{bingo_cui2023holistic} & arXiv'23 Nov. & Unknown & 370 & Gen & Human Assessment & \xmark & \xmark & \xmark & Model Bias \\
        AMBER~\cite{amber} & arXiv'23 Nov. & Web & 15,202 & Dis \& Gen & AMBER Score & \cmark & \cmark & \cmark & \xmark \\
        RAH-Bench~\cite{RAH_bench} & arXiv'23 Nov. & MSCOCO~\cite{mscoco} & 3,000 & Dis & False Positive Rate & \cmark & \cmark & \cmark & \xmark \\
        HallusionBench~\cite{liu2023hallusionbench} & CVPR'24 & Unknown & 1,129 & Gen & LLM Assessment & \xmark & \xmark & \xmark & Model Diagnose \\
        CCEval~\cite{halle_switch} & arXiv'23 Dec. & Visual-Genome~\cite{visual_genome} & 100 & Gen & LLM-based CHAIR & \cmark & \xmark & \xmark & \xmark \\
        MERLIM~\cite{villa2023behind} & arXiv'23 Dec. & MSCOCO~\cite{mscoco} & 31,373 & Dis & Accuracy & \cmark & \xmark & \cmark & Obj. Counting \\
        FGHE~\cite{fghe_wang2023mitigating} & MMM'24 & MSCOCO~\cite{mscoco} & 200 & Dis & Acc/P/R/F & \cmark & \cmark & \cmark & Obj. Behavior \\
        MOCHa~\cite{ben2023mocha_openchair} & EMNLP'24 & Synthetic & 2,000 & Gen & OpenCHAIR~\cite{ben2023mocha_openchair} & \cmark & \cmark & \xmark & \xmark \\

        CorrelationQA~\cite{correlation_qa} & arXiv'24 Feb. & Synthetic & 7,308 & Dis & Acc/AccDrop & \xmark & \xmark & \xmark & Model Bias \\
        VQAv2-IDK~\cite{vqav2_idk} & ICASSP'24 & VQAv2~\cite{vqav2} & 6,624 & Dis & Acc & \xmark & \xmark & \xmark & IK~\cite{vqav2_idk} \\
        MHaluBench~\cite{easy_detect} & ACL'24 & MSCOCO~\cite{mscoco} & 1,860 & Gen & Acc/P/R/F & \cmark & \cmark & \xmark & T2I \\

        VHTest~\cite{vhtest} & ACL'24 & MSCOCO~\cite{mscoco} & 1,200 & Dis \& Gen & Acc & \cmark & \cmark & \xmark & \cmark \\

        Hal-Eavl~\cite{hal_eval} & MM'24 & \makecell[l]{MSCOCO~\cite{mscoco} \& \\ LAION~\cite{schuhmann2022laion}} & 10,000 & Dis \& Gen & \makecell[c]{Acc/P/R/F \& \\ LLM Assessment} & \cmark & \cmark & \cmark & Obj. Event \\

        PhD~\cite{liu2024phd} & arXiv'24 Mar. & TDIUC~\cite{tdiuc} \& AIGC & 102,564 & Dis & PhD Index & \cmark & \cmark & \cmark & Sentiment \\
        THRONE~\cite{kaul2024throne} & CVPR'24 & MSCOCO~\cite{mscoco} & 5,000, & Gen & P/R/F & \cmark & \xmark & \xmark & \xmark \\
        BEAF~\cite{ye2024beaf} & ECCV'24 & MSCOCO~\cite{mscoco} & 26,118 & Dis & TU/IG/SB/ID & \cmark & \xmark & \xmark & \xmark \\
        ROPE~\cite{rope} & NeurIPS'24 & \makecell[l]{MSCOCO~\cite{mscoco} \& \\ ADE20k~\cite{ade20k}} & 5,000 & Dis & Acc & \cmark & \xmark & \xmark & Multi Obj. \\
        LongHalQA~\cite{qiu2024longhalqa} & arXiv'24 Oct. &  \makecell[l]{VisualGenome~\cite{visual_genome} \& \\ Objects365~\cite{shao2019objects365}} & 6,485 & Dis \& Gen & Acc & \cmark & \cmark & \cmark & Obj. Counting \\
        Reefknot~\cite{zheng2024reefknot} & arXiv'24 Dec. & VisualGenome~\cite{visual_genome} & 21,880 & Dis \& Gen & R-score & \xmark & \xmark & \cmark & \xmark \\
        
        \bottomrule
        
    \end{tabular}}
    \label{tab:summary_bench_metrics}
\end{table*}

In this section, we present a comprehensive overview of existing hallucination metrics and benchmarks, which are designed to assess the extent of hallucinations generated by existing cutting-edge MLLMs.
Currently, the primary focus of these benchmarks is on evaluating the object hallucination of MLLM-generated content.
Tab.~\ref{tab:summary_bench_metrics} illustrates a summary of related benchmarks.

\textbf{CHAIR~\cite{chair}}.
As one of the early works, the metric of CHAIR was proposed to evaluate object hallucination in the traditional image captioning task.
This is achieved by computing what proportion of words generated are actually in the image according to the ground truth sentences and object segmentations.
The computation of the CHAIR metric is straightforward and easy to understand.
The metric has two variants: per-instance (denoted as $\text{CHAIR}_i$) and per-sentence (denoted as $\text{CHAIR}_s$):
\begin{equation*}
    \text{CHAIR}_i = \frac{|\{\text{hallucinated objects}\}|}{|\{\text{all objects mentioned}\}|},
\end{equation*}
\begin{equation*}
    \text{CHAIR}_s = \frac{|\{\text{sentences with hallucinated object}\}|}{|\{\text{all sentences}\}|}.
\end{equation*}
In the paper of CHAIR~\cite{chair}, the range of objects is restricted to the 80 MSCOCO objects.
Sentence tokenization and synonyms mapping are applied to determine whether a generated sentence contains hallucinated objects.
Ground-truth caption and object segmentations both serve as ground-truth objects in the computation.
In the MLLM era, this metric is still widely used for assessing the response of MLLMs.

\textbf{POPE~\cite{pope}}.
When used in MLLMs, the work of \cite{pope} argues that the CHAIR metric can be affected by the instruction designs and the length of generated captions.
Therefore, it proposes a new evaluation metric as well as a benchmark, called Pooling-based Object Probing Evaluation (POPE).
The basic idea is to convert the evaluation of hallucination into a binary classification task by prompting MLLMs with simple \textit{Yes}-or-\textit{No} short questions about the probing objects (\textit{e.g.}, Is there a \textit{car} in the image?)
Compared to CHAIR, POPE offers increased stability and flexibility.
Based on this metric design, it further proposed an evaluation benchmark, drawing 500 images from the MSCOCO dataset.
The questions in the benchmark consist of both positive and negative questions.
The positive questions are formed based on the ground-truth objects, while the negative questions are built from sampling nonexistent objects.
The benchmark is divided into three subsets according to different negative sampling strategy: random, popular, and adversarial.
Popular and adversarial sampling are specifically designed to assess frequently appeared objects and object co-occurrence.
As an early representative work, POPE serves as a foundation of object hallucination evaluation.

\textbf{MME~\cite{mllm_survey_mme}}.
MME is a comprehensive evaluation benchmark for MLLMs.
It covers the examination of perception and cognition abilities, encompassing 14 subtasks.
Regarding object hallucination, there are four popular object related subtasks in its perception evaluation, including object existence, count, position, color.
Similar to POPE, these tasks are formulated as \textit{Yes}-or-\textit{No} tasks.

\textbf{CIEM~\cite{hu2023ciem}}
CIEM is a benchmark to evaluate hallucination of MLLMs.
Unlike previous works utilize human annotated objects, CIEM is generated using an automatic pipeline.
The pipeline takes the text description of a specific image as input and utilize advanced LLMs to generate QA pairs.
Although the LLM-based data generation pipeline is not completely reliable, empirical result shows that the generated data has low error rate, around 5\%.

\textbf{MMHal-Bench~\cite{llava_rlhf_sun2023aligning}}
Comprising 96 image-question pairs, ranging in 8 question categories $\times$ 12 object topics, MMHal-Bench is a dedicated benchmark for evaluating hallucination in MLLMs.
The 8 question categories cover various types of hallucination, including object attributes, counting, spatial relations, etc.
During the evaluation of MMHal-Bench, the GPT-4 model is employed to analyze and rate the responses.

\textbf{GAVIE~\cite{LRV_instruction}}
GPT4-Assisted Visual Instruction Evaluation (GAVIE) is proposed to assess the LMM output in two different aspects: \textit{Relevancy} to evaluate the instruction-following performance and \textit{Accuracy} to measure the visual hallucination in the LMM output.
It comprises a benchmark with 1,000 samples and an evaluation approach.
GAVIE evaluates the output of MLLMs in an open-ended manner and does not require human-annotated ground-truth answers.
The core idea is to ask the advanced GPT-4 to work as a smart teacher and score the answer by taking image content, human instruction, and model response as input.

\textbf{NOPE~\cite{NOPE}}
This paper proposes to establish a distinction between object hallucination and incorrectness.
a) Object hallucination refers to a phenomenon in VQA where a VL model's response includes a non-existent object, despite the ground truth answer being a negative indefinite pronoun (e.g., "none", "no one", etc). This is denoted as \textsc{NegP}.
b) Incorrectness occurs when a VL model fails to accurately respond to a question with a ground truth answer that is anything other than \textsc{NegP}, denoted as \textsc{Others}.
This paper argues that the existing VQA datasets have a significantly imbalanced distribution, containing too little \textsc{NegP} data.
Therefore, NOPE (Negative Object Presence Evaluation) is proposed in this paper to complement the absent \textsc{NegP} data.
During evaluation, traditional metrics, including Accuracy and METEOR, are employed.

\textbf{HaELM~\cite{haelm}}
Most LLM-based evaluation benchmarks employ advanced ChatGPT or GPT-4 models to assess the quality of the MLLM response.
In contrast, the work of Hallucination Evaluation based on Large Language Models (HaELM) proposes to train a specialized LLM for hallucination detection.
It collects a set of hallucination data generated by a wide range of MLLMs, simulates data using ChatGPT, and trains an LLM based on LLaMA~\cite{touvron2023llama}.
After that, the HaELM model becomes proficient in hallucination evaluation, leveraging reference descriptions of images as the basis of assessment.

\textbf{FaithScore~\cite{jing2023faithscore}}
Considering the natural forms of interaction between humans and MLLMs, FaithScore aims to evaluate free-form responses to open-ended questions.
Different from LLM-based overall assessment, FaithScore designs an automatic pipeline to decompose the response, evaluate, and analyze the elements in detail.
Specifically, it includes three steps: descriptive sub-sentence identification, atomic fact generation, and fact verification.
The evaluation metric involves fine-grained object hallucination categories, including entity, count, color, relation, and other attributes.
The final computation of FaithScore is the ratio of hallucinated content.

\textbf{Bingo~\cite{bingo_cui2023holistic}}
Bingo (Bias and Interference Challenges in Visual Language Models) is a benchmark specifically designed for assessing and analyzing the limitations of current popular MLLMs, such as GPT-4V~\cite{gpt4v_card}.
It comprises 190 failure instances, along with 131 success instances as a comparison.
This benchmark reveals that state-of-the-art MLLMs show the phenomenon of bias and interference.
Bias refers to the model's susceptibility to generating hallucinatory outputs on specific types of examples, such as OCR bias, region bias, etc.
Interference refers to scenarios in which the judgment of the model can be disrupted, making it more susceptible to hallucination.
Due to the small amount of data in this benchmark, the assessment and analysis are mostly conducted by humans.

\textbf{AMBER~\cite{amber}}
Upon the application and evaluation of MLLMs, the tasks can be roughly divided into generative tasks and discriminative tasks.
For generative tasks, this paper argues that most existing works rely on additional LLMs, suffering from computational cost.
As for discriminative tasks, the most popular evaluation suite is POPE~\cite{pope}.
However, POPE lacks fine-grained hallucination types such as attributes and relations.
AMBER (An LLM-free Multi-dimensional Benchmark) is proposed to support the evaluation of generative tasks and discriminative tasks, including object existence hallucination, attribute hallucination, and relation hallucination.
It further combines the \textbf{CHAIR}~\cite{chair} metric in generative tasks and \textbf{F1} in discriminative tasks to form the AMBER Score as follows:
\begin{equation}
    \textbf{AMBER Score} = Avg(1-\textbf{CHAIR}, \textbf{F1}).
\end{equation}

\textbf{RAH-Bench~\cite{RAH_bench}}
Relation-Associated Hallucination Benchmark (RAH-Bench) can be regarded as an upgraded version of POPE, containing 3,000 yes-or-no questions with their corresponding images.
Different from POPE, RAH-Bench further divides the negative questions into three subsets.
Each subset contains 500 questions with misleading statements in the different aspects, including: 1) categorical hallucination, 2) attribute hallucination, 3) relation hallucination.

\textbf{HallusionBench~\cite{liu2023hallusionbench}}
To diagnose and analyze the potential failure modes of MLLMs, HallusionBench evaluates hallucination from a different perspective.
It consists of 455 visual-question control pairs, with 346 different figures and a total of 1129 questions covering diverse topics and formats.
The questions are divided into two categories: \textit{Visual Dependent} and \textit{Visual Supplement}.
The \textit{Visual Dependent} questions are defined as questions that do not have an affirmative answer without the visual context.
This setting aims to evaluate visual commonsense knowledge and visual reasoning skills.
The \textit{Visual Supplement} questions can be answered without the visual input; the visual component merely provides supplemental information or corrections.
This setting is designed to evaluate visual reasoning ability and the balance between parametric memory (language prior) and image context.
This division provides a new perspective for understanding and diagnosing MLLMs.

\textbf{CCEval~\cite{halle_switch}}
CCEval focuses on the hallucination evaluation of detailed captions.
Traditional caption-based evaluation benchmarks and metrics, like CHAIR, are known to favor short captions.
However, short captions often lack detail and contain less information.
To address this issue, CCEval randomly samples 100 images from Visual Genome to form a benchmark.
In evaluation, GPT-4 is utilized to parse the captions generated by MLLMs and extract objects.
Additionally, this work introduces the "coverage" metric on top of CHAIR to ensure that the captions are detailed enough.
This metric computes the ratio of objects in the caption that match the ground truth to the total number of ground truth objects.
It additionally records the average number of objects as well as the average length of captions as auxiliary metric.
Compared with CHAIR, CCEval employs more diverse objects, as reflected in the source of ground truth (Visual Genome vs. COCO) and caption parsing (GPT-4 vs. rule-based tool).

\textbf{MERLIM~\cite{villa2023behind}}
MERLIM (\textbf{M}ulti-modal \textbf{E}valuation benchma\textbf{R}k for \textbf{L}arge \textbf{I}mage-language \textbf{M}odels) is a test-bed aimed at empirically evaluating MLLMs on core computer vision tasks, including object recognition, instance counting, and identifying object-to-object relationships.
MERLIM contains over 279K image-question pairs, and has a strong focus on detecting cross-modal hallucinations.
Interestingly, when organizing the data, a set of edited images is intentionally added.
Based on the original image, an inpainting strategy is employed to remove one object instance in the image.
With this original-edited image pair, one can compare the output of the target MLLM and identify the hallucinated objects that lack visual grounding.

\textbf{FGHE~\cite{fghe_wang2023mitigating}}
Fine-Grained Object Hallucination Evaluation (FGHE) follows a binary classification approach similar to POPE to evaluate MLLMs.
However, unlike POPE, FGHE requires a different set of binary questions to measure fine-grained hallucination.
The FGHE dataset consists of 50 images and 200 binary questions divided into three categories: (a) multiple-object questions, which verify the relationships between multiple objects in the image; (b) attribute questions, which verify attributes of objects in the image; and (c) behavior questions, which verify behaviors or objects in the image.
The questions are manually defined by human annotators on a subset of 50 images from the validation set of the MSCOCO dataset.
Similar to POPE, the Accuracy, Precision, Recall, and F1 score are employed as the evaluation metrics.

\textbf{OpenCHAIR~\cite{ben2023mocha_openchair}}
The traditional CHAIR metric relies on the closed list of 80 objects in the MS-COCO dataset, limiting its application.
To measure object hallucination in the open-vocabulary settings, \textit{OpenCHAIR} expands CHAIR by relaxing the strong reliance on the closed vocabulary.
The 'open-vocabulary' manifests in two ways.
Firstly, when building the benchmark, it organizes a dataset consisting of synthetic images with corresponding captions, which include diverse, open-vocabulary objects using a text-to-image diffusion model.
Secondly, during computing the metric, CHAIR checks if words or their synonyms (as given by fixed vocabulary lists) are found in ground-truth annotations.
In contrast, OpenCHAIR extracts concrete objects from a predicted caption and identifies hallucinated objects from this list by querying an LLM.
Similar to CHAIR, the final metric computation is based on the hallucination rate.

\textbf{Hal-Eval~\cite{hal_eval}}
The work of Hal-Eval~\cite{hal_eval} identifies another type of object hallucination: event hallucination.
This type of hallucination fabricates a fictional target and constructing an entire narrative around it, including its attributes, relationships, and actions.
This effort further completes the definition of hallucination types.
In addition, this work proposes an evaluation benchmark, which encompasses both discriminative and generative evaluation methods.
This is achieved by collecting two evaluation subsets, each tailored to the discriminative and generative evaluation methods, respectively.

\textbf{CorrelationQA~\cite{correlation_qa}}
CorrelationQA is a dedicated benchmark to quantify the effect of hallucination induced by the spurious visual input.
This type of hallucination usually occurs when providing the MLLM with images that are highly relevant but inconsistent with the answers, causing MLLMs to suffer from hallucination.
Such visual inputs are defined as 'spurious visual inputs'.
This benchmark reveals that most of mainstream MLLMs, including GPT-4V, suffer from hallucination when presented with such spurious visual inputs.
This phenomenon indicates that an image can induce MLLMs to instinctively focus on visual content, resulting in responses that are predominantly based on visual information without proper reasoning and thinking.

\textbf{VQAv2-IDK~\cite{vqav2_idk}}
It has been widely discussed that in the binary QA scenario, MLLMs generally have a bias on answering 'Yes-or-No,' leading to hallucination.
In a more detailed question and answer scenario, MLLMs generally tend to respond to the user's question plausibly, even if the desired answer is 'I don't know'.
The concept is defined as \textit{'I Know (IK)' hallucination} in the work of~\cite{vqav2_idk}.
Accordingly, a new benchmark, VQAv2-IDK, is proposed to specifically evaluate this type of hallucination.
VQAv2-IDK is a subset of VQAv2 comprising unanswerable image-question pairs as determined by human annotators.
In this benchmark, 'I Know (IK)' hallucination has been further categorized into four types:
\begin{itemize}
    \item Unanswerable: no one can know.
    \item Don't know: human may not know, but robot might.
    \item False questions: refers non-existing.
    \item Not sure: ambiguous to answer.
\end{itemize}
This benchmark opens a new track for the study of hallucination in MLLMs.

\textbf{MHaluBench~\cite{easy_detect}}
This benchmark does not aim to evaluate the MLLMs themselves.
Instead, it is intentionally designed to evaluate the hallucination detection tools of MLLMs, \textit{i.e.}, judge whether a tool can successfully detect the hallucination produced by an MLLM.
Thus, the benchmark consists of hallucinatory examples.
Specifically, the benchmark unifies image-to-text tasks and the text-to-image tasks into one evaluation suite: cross-modal consistency checking.
The hallucinatory examples are generated using leading MLLMs and image generation models, such as LLaVA~\cite{llava}, MiniGPT-4~\cite{minigpt}, DALL-E2~\cite{dalle2}, and DALL-E3~\cite{dalle3}.
During evaluation, the benchmark can be used to compare different hallucination detection methods based on their performance.
So far, there are not many dedicated hallucination detection methods.
This work serves as a basis for this direction.

\textbf{VHTest~\cite{vhtest}}
VHTest categorizes visual properties of objects in an image into 1) individual properties, such as existence, shape, color, orientation, and OCR; and 2) group properties, which emerge from comparisons across multiple objects, such as relative size, relative position, and counting.
Based on such categorization, the authors further defined 8 visual hallucination modes, providing a very detailed evaluation of hallucination in MLLMs.
Furthermore, the collected 1,200 evaluation instances are divided into two versions: "open-ended question" (OEQ) and "yes/no question" (YNQ).
Such design enables this benchmark to evaluate both generative and discriminative tasks.

\textbf{THRONE~\cite{kaul2024throne}}
THRONE categorizes hallucinations in generative and discriminative tasks as Type I and Type II hallucination.
It argues that most existing benchmarks focus on Type II and result does not translate into Type I.
Regarding evaluating Type I, the main drawback of existing works is the reliance on close-source GPT model.
Thus, this works proposes an evaluation framework based on MSCOCO dataset, which 1) evaluates object hallucination in free-form output; 2) utilizes open-source LLM instead of close-source ones.

\textbf{BEAF~\cite{ye2024beaf}}
BEAF curates a new evaluation dataset, called the BEfore-AFter hallucination dataset (BEAF), and introduces new metrics: True Understanding(TU), IGnorance (IG), StuBbornness (SB), and InDecision (ID).
Different from prior works that focus only on constructing questions and answers, the key idea of this benchmark is to manipulate visual scene information by image editing models and to design the metrics based on scene changes.
This allows clearly assessment on whether VLMs correctly understand a given scene by observing the ability to perceive changes.

\textbf{Comparison of mainstream models}
We compare the mainstream MLLMs on some representative benchmarks, providing a holistic overview of their performance from different dimensions.
The results are shown in Table~\ref{tab:comp_gen} for generative tasks and Table~\ref{tab:comp_dis} for discriminative tasks.
We observe that the MLLMs' performance is not always consistent across different benchmarks.
It indicates that different benchmarks have different evaluation dimensions and emphases.

\begin{table*}[ht!]
    \centering
    \renewcommand*{\arraystretch}{1}
    \caption{
    Comparison of mainstream MLLMs on \textbf{generative} benchmarks.
    The numbers come from the original papers of these benchmarks.
    }
    \setlength\tabcolsep{4pt}
    \resizebox{\linewidth}{!}{
    \begin{tabular}{@{}lcccccccc@{}}
        \toprule
        Model & 
        LLM Size &
        \makecell[c]{CHAIR\\ (On AMBER) $\downarrow$} &
        \makecell[c]{AMBER\\ Score $\uparrow$} &
        \makecell[c]{HallusionBench\\ All-Acc $\uparrow$} &
        \makecell[c]{FaithScore\\ (LLaVA-1k) $\uparrow$} &
        \makecell[c]{FaithScore\\ (COCO-Cap) $\uparrow$} &
        \makecell[c]{Hal-Eval \\ In-domain\\ Gen. Acc $\uparrow$} &
        \makecell[c]{Hal-Eval \\ Out-of-domain\\ Gen. Acc $\uparrow$} \\
        \midrule
        mPLUG-Owl~\cite{mplug}                          & 7B    & 23.1  & 54.1  & 43.93     & 0.7167    & 0.8546    & 27.3  & 29.5              \\
        Multimodal-GPT~\cite{gong2023multimodalgpt}     & 7B    & -     & -     & -         & 0.5335    & 0.5440    & -     & -                 \\
        InstructBLIP~\cite{instructblip}                & 7B    & 10.3  & 86.2  & 45.26     & 0.8091    & 0.9392    & 35.5  & 41.3              \\
        GPT-4V~\cite{gpt4v_card}                        & -     & 4.3   & 92.7  & 65.28     & -         & -         & -     & -                 \\
        LLaVA (7B)~\cite{llava}                         & 7B    & 13.5  & 69.3  & -         & -         & -         & 23.3  & 26.3              \\
        LLaVA (13B)~\cite{llava}                        & 13B   & -     & -     & -         & 0.8360    & 0.8729    & -     & -              \\
        MiniGPT-4 (7B)~\cite{minigpt}                   & 7B    & -     & -     & 35.78     & 0.5713    & 0.6359    & 61.4  & 50.1              \\
        MiniGPT-4 (13B)~\cite{minigpt}                  & 13B   & 15.9  & 76.7  & -         & -         & -         & -  & -              \\
        mPLUG-Owl2~\cite{mplug_2}                       & 7B    & 10.6  & 84.0  & 47.30     & -         & -         & -     & -                 \\
        LLaVA-1.5 (7B)~\cite{llava15}                   & 7B    & 8.6   & 82.9  & -         & -         & -         & 44.6  & 46.4              \\
        LLaVA-1.5 (13B)~\cite{llava15}                  & 13B   & -     & -     & 46.94     & 0.8566    & 0.9425    & -     & -              \\
        CogVLM~\cite{wang2023cogvlm}                    & 7B    & 7.9   & 86.1  & -         & -         & -         & -     & -                 \\
        Qwen-VL-Chat~\cite{bai2023qwen}                 & 7B    & -     & -     & 39.15     & -         & -         & -     & -                 \\
        Open-Flamingo~\cite{awadalla2023openflamingo}   & 9B    & -     & -     & 38.44     & -         & -         & -     & -                 \\
        LRV-Instruction~\cite{LRV_instruction}          & -     & -     & -     & 42.78     & -         & -         & -     & -                 \\
        \bottomrule
    \end{tabular}}
    \label{tab:comp_gen}
\end{table*}

\begin{table*}[ht!]
    \centering
    \renewcommand*{\arraystretch}{1}
    \caption{
    Comparison of mainstream MLLMs on \textbf{discriminative} benchmarks.
    The numbers come from the original papers of these benchmarks.
    }
    \setlength\tabcolsep{4pt}
    \resizebox{\linewidth}{!}{
    \begin{tabular}{@{}lccccccccccccccc@{}}
        \toprule
        Model & 
        \makecell{LLM \\ Size} &
        \makecell{MME \\ Existence \\ Score $\uparrow$} &
        \makecell{MME \\ Count\\ Score $\uparrow$} &
        \makecell{MME \\ Position \\ Score $\uparrow$} &
        \makecell{MME \\ Color \\ Score $\uparrow$} &
        \makecell{POPE \\ Random \\ F1-Score $\uparrow$} &
        \makecell{POPE \\ Random \\ F1-Score $\uparrow$} &
        \makecell{POPE \\ Adversarial \\ F1-Score $\uparrow$} &
        \makecell{RAH-Bench\\ F1 Score $\uparrow$} &
        \makecell{AMBER\\ Dis.\\ F1-Score $\uparrow$} &
        \makecell{AMBER\\ Score $\uparrow$} &
        \makecell{Hal-Eval \\ In-domain\\ Event. F1 $\uparrow$} &
        \makecell{Hal-Eval \\ Out-of-domain\\ Event. F1 $\uparrow$} \\
        \midrule
        mPLUG-Owl~\cite{mplug}                      & 7B    & 120.00    & 50.00        & 50.00        & 55.00        & 68.06 & 66.79 & 66.82 & 69.3 & 31.2 & 54.1 & 47 & 46.6 \\
        ImageBind-LLM~\cite{han2023imagebindllm}    & 7B    & 128.33    & 60.00        & 46.67     & 73.33     & - & - & - & - & - & - & - & - \\
        InstructBLIP~\cite{instructblip} (7B)       & 7B    & -         & -         & -         & -         & - & - & - & 89.1 & 82.6 & 86.2 & 66.2 & 66.6 \\
        InstructBLIP~\cite{instructblip} (13B)      & 13B   & 185.00       & 143.33    & 66.67     & 153.33    & 89.29 & 83.45 & 78.45 & 84.7 & - & - & - & - \\
        VisualGLM-6B~\cite{du2022glm}               & 6B    & 85.00        & 50.00        & 48.33     & 55.00        & - & - & - & - & - & - & - & - \\
        Multimodal-GPT~\cite{gong2023multimodalgpt} & 7B    & 61.67     & 55.00        & 58.33     & 68.33     & 66.68 & 66.67 & 66.67 & - & - & - & - & - \\
        PandaGPT~\cite{su2023pandagpt}              & 7B    & 70.00        & 50.00        & 50.00        & 50.00        & - & - & - & - & - & - & - & - \\
        LaVIN~\cite{lavin}                          & 13B   & 185.00       & 88.33     & 63.33     & 75.00        & - & - & - & - & - & - & - & - \\
        Cheetor~\cite{cheetor}                      & 7B    & 180.00       & 96.67     & 80.00        & 116.67    & - & - & - & - & - & - & - & - \\
        GPT-4V~\cite{gpt4v_card}                    & -     & 190.00       & 160.00       & 95.00        & 150.00       & - & - & - & - & 89.6 & 92.7 & - & - \\
        LLaVA~\cite{llava} (7B)                     & 7B    & -         & -         & -         & -         & - & - & - & 73.3 & 32.0 & 69.3 & 35.1 & 14.0 \\
        LLaVA~\cite{llava} (13B)                    & 13B   & 185.00       & 155.00       & 133.33    & 170.00       & 68.65 & 67.72 & 66.98 & 71.8 & - & - & - & - \\
        LRV-Instruction~\cite{LRV_instruction}      & 7B    & 165.00       & 111.67    & 86.67     & 165.00       & - & - & - & - & - & - & - & - \\
        Lynx~\cite{lynx}                            & 7B    & 195.00       & 151.67    & 90.00        & 170.00       & - & - & - & - & - & - & - & - \\
        MMICL~\cite{zhao2023mmicl}                  & 11B   & 170.00       & 160.00       & 81.67     & 156.67    & - & - & - & - & - & - & - & - \\
        Muffin~\cite{yu2023muffin}                  & 13B   & 195.00       & 163.33    & 66.67     & 165.00       & - & - & - & - & - & - & - & - \\
        Otter~\cite{li2023otter}                    & 7B    & 195.00       & 88.33     & 86.67     & 113.33    & - & - & - & - & - & - & - & - \\
        Qwen-VL-Chat~\cite{bai2023qwen}             & 7B    & 158.33    & 150.00       & 128.33    & 170.00       & - & - & - & - & - & - & - & - \\
        SPHINX~\cite{lin2023sphinx}                 & 13B   & 195.00       & 160.00       & 153.33    & 160.00       & - & - & - & - & - & - & - & - \\
        VPGTrans~\cite{zhang2024vpgtrans}           & 7B    & 70.00        & 85.00        & 63.33     & 73.33     & - & - & - & - & - & - & - & - \\
        BLIVA~\cite{hu2023bliva}                    & 11B   & 180.00         & 138.33    & 81.67     & 180.00       & - & - & - & - & - & - & - & - \\
        InfMLLM~\cite{zhou2023infmllm}              & 13B   & 195.00       & 145.00       & 170.00       & 195.00       & - & - & - & - & - & - & - & - \\
        LLaMA-Adapter V2~\cite{llama_ada_v2}        & 7B    & 185.00       & 133.33    & 56.67     & 118.33    & - & - & - & - & - & - & - & - \\
        MiniGPT-4~\cite{minigpt}                    & 13B   & 68.33     & 55.00        & 43.33     & 75.00      & 78.86 & 72.21 & 71.37 & - & 69.3 & 76.7 & 48.2 & 53.0 \\
        mPLUG-Owl2~\cite{mplug_2}                   & 7B    & 185.00       & 155.00       & 88.33     & 150.00       & - & - & - & - & 78.5 & 84.0 & - & - \\
        LLaVA-1.5~\cite{llava}                      & 7B    & -         & -         & -         & -         & 83.44 & 82.08 & 79.26 & - & 74.4 & 82.9 & 48.9 & 34.2 \\
        CogVLM~\cite{wang2023cogvlm}                & 7B    & 195.00       & 165.00       & 103.33    & 160.00       & - & - & - & - & 80 & 86.1 & - & - \\

        Intern-VL2~\cite{chen2024internvl} & 7B & - & - & - & - & 82.60 & 81.64 & 80.70 & - & - & - & - & - \\
        DeepSeek-VL2~\cite{wu2024deepseekvl2} & 7B & - & - & - & - & 87.64 & 85.86 & 85.55 & - & - & - & - & - \\
        Qwen-VL2~\cite{wang2024qwen2} & 7B & - & - & - & - & 88.99 & 87.68 & 86.02 & - & - & - & - & - \\
        LLaVA-One-Vision~\cite{li2024llavaonevision} & 7B & - & - & - & - & 88.51 & 87.33 & 86.27 & - & - & - & - & - \\

        OPERA(LLaVA-1.5)~\cite{huang2023opera} & 7B & 180.67 & 133.33 & 123.33 & 155.00 & 89.95 & 86.88 & 81.77 & - & - & - & - & - \\
        VCD(LLaVA-1.5)~\cite{VCD} & 7B & 186.67 & 125.56 & 128.89 & 139.45 & 87.15 & 83.94 & 79.47 & - & - & - & - & - \\
        M3ID(LLaVA-1.5)~\cite{M3ID} & 7B & 186.67 & 128.33 & 131.67 & 151.67 & 87.52 & 84.95 & 80.22 & - & - & - & - & - \\
        RITUAL(LLaVA-1.5)~\cite{woo2024ritual} & 7B & 187.50 & 139.58 & 125 & 164.17 & 88.81 & 86.17 & 80.54 \\
        DeGF(LLaVA-1.5)~\cite{DeGF} & 7B & 188.33 & 150.00 & 133.89 & 172.22 & 88.74 & 86.28 & 81.94 & - & - & - & - & - \\
        SID(LLaVA-1.5)~\cite{SID} & 7B & 190.00 & 148.33 & 128.33 & 175.00 & 89.00 & 85.01 & 81.28 & - & - & - & - & - \\
        CCA(LLaVA-1.5)~\cite{CCA} & 7B & 190.00 & 148.33 & 128.33 & 175.00 & 89.05 & 86.02 & 83.82 & - & - & - & - & - \\
        DAC(LLaVA-1.5)~\cite{VAP} & 7B & 195.00 & 158.33 & 133.33 & 170.00 & 90.60 & 89.10 & 84.42 & - & - & - & - & - \\

        \bottomrule
    \end{tabular}}
    \label{tab:comp_dis}
\end{table*}

\section{Hallucination Mitigation}
\label{sec:mitigating}

In this section, we present a comprehensive review of contemporary methods aimed at mitigating hallucinations in MLLMs.
Based on the properties and perspectives of these methods, we systematically categorize them into four groups.
Specifically, we investigate approaches addressing hallucination from Data, Model, Training, and Inference.

\subsection{Data}
\label{sec:mitigating_data}
As discussed in the section on hallucination causes~\ref{sec:causes}, data is one of the primary factors inducing hallucination in MLLMs.
For mitigating hallucination, recent works make attempts on data, including introducing negative data~\cite{LRV_instruction}, counterfactual data~\cite{yu2023hallucidoctor,chen2025perturbollava}, reasoning data~\cite{zhang2024reflective}, and reducing noise and errors in existing dataset~\cite{fghe_wang2023mitigating, eos_token}.

\subsubsection{Negative Data}
\textbf{LRV-Instruction~\cite{LRV_instruction}}
LRV-Instruction is proposed to address the issue that existing instruction tuning data primarily focus on positive instruction samples, leading the model to consistently answer 'Yes'.
LRV-Instruction is designed to include both positive and negative instructions for more robust visual instruction tuning, where the negative instructions include:
1) 'Nonexistent Object Manipulation': introducing nonexistent objects, activities, attributes, and interactions;
2) 'Existent Object Manipulation': manipulating existent objects with inconsistent attributes;
3) 'Knowledge Manipulation': manipulating knowledge in instructions.

\subsubsection{Counterfactual Data}
\textbf{HalluciDoctor~\cite{yu2023hallucidoctor}}
This paper addresses the object hallucination problem in MLLMs by calibrating the instruction-tuning dataset.
The calibration is conducted from two perspectives.
Firstly, it develops a hallucination detection pipeline via consistency cross-checking of multiple MLLMs.
Based on the detection result, the hallucinated content can be eliminated.
Secondly, this work observes that long-tail distribution and object co-occurrence in the training data are two primary factors of hallucination.
Thus, a counterfactual visual instruction generation strategy is proposed to expand the dataset.
Using the proposed methods, the instruction tuning data can be balanced and experience reduced hallucination.
MLLMs trained on the calibrated dataset are shown to be less prone to hallucination.

Recently, PerturboLLaVA~\cite{chen2025perturbollava} explores a similar strategy.
It reduces the model’s heavy dependence on the language prior by incorporating adversarially perturbed text during training.
Specifically, it introduce carefully designed perturbations that aligns with the general knowledge but conflict with the visual content, intentionally misleading the model based on its language prior.
This perturbative training enforces the model to scrutinize the image content when predicting every token, rather than hallucinating contents from the text hints.
Essentially, this method adjusts the model’s conditional distribution to depend more heavily on the image and less on the perturbation text, which leads to more robust multimodal capability.

\subsubsection{Reasoning Data}
\textbf{REVERIE~\cite{zhang2024reflective}}
This paper argues that without intermediate reasoning steps, models may establish superficial shortcuts between instructions and responses, failing to internalize the inherent reasoning logic. 
To address this challenge, it propose reflective instruction tuning, which integrates rationale learning into visual instruction tuning.
Unlike previous methods that learning from responses only, this approach entails the model predicting rationales justifying why responses are correct or incorrect.
This instruction tuning is supported by a specially curated dataset REVERIE.
Each instruction is meticulously annotated with a corresponding pair of correct and confusing responses, alongside comprehensive rationales elucidating the justification behind the correctness or erroneousness of each response.

\subsubsection{Clean Data}
\textbf{ReCaption~\cite{fghe_wang2023mitigating}}
This work proposes a framework called ReCaption to rewrite the text captions of existing image-text pairs in datasets.
The framework comprises two steps: 1) keyword extraction, which extracts verbs, nouns, and adjectives from the caption; and 2) caption generation, which employs an LLM to generate sentences based on the extracted keywords.
Ultimately, the framework produces a set of high-quality image-caption pairs.
Experiment results show that the model trained on the rewritten caption dataset has higher accuracy on certain benchmarks, such as the POPE benchmark~\cite{pope}.
Despite the performance improvement, the question of why rewritten captions can reduce hallucination remains an open problem.

\textbf{EOS Decision~\cite{eos_token}}
Previous work~\cite{LURE_zhou2023analyzing} provides an observation that hallucination tends to occur with objects positioned later in the generated descriptions.
Intuitively, an ideal scenario is that the MLLM can terminate the generation process in a timely manner.
This idea is thoroughly explored in the work of~\cite{eos_token} from the perspective of end-of-sequence (EOS) decision.
The key insight is that the training data may exceed the perception limit of the MLLM.
When trained with such data, the model may attempt to fit the detail level and length distribution of ground truth captions.
However, it may risk expressing details that it cannot discern from the image, and therefore exhibit hallucinations.
Thus, the authors explored approaches to enhance the model's end-of-sequence (EOS) decision-making process, ensuring timely termination when it reaches the perception limit.
Regarding data, this work proposes a data filtering strategy to eliminate harmful training data that could impair the model's ability to end sequences.

\subsection{Model}
\label{sec:mitigating_model}

\subsubsection{Scale-up Resolution}
Enhancing the perception ability of MLLMs has been shown to improve their overall performance and reduce hallucination~\cite{llava,llava15,halle_switch,chen2024internvl}.
One important update when upgrading from LLaVA~\cite{llava} to LLaVA-1.5~\cite{llava15} is to scale up the CLIP ViT vision encoder from CLIP-ViT-L-224 to CLIP-ViT-L-336, resulting in considerable performance improvement.
Qwen-VL~\cite{bai2023qwen} has shown the effectiveness of gradually enlarging image resolution from $224\times 224$ to $448\times 448$.
InternVL~\cite{chen2024internvl} scales up the vision encoder to 6 billion parameters, enabling processing of high-resolution images.
Regarding hallucination, HallE-Switch~\cite{halle_switch} has investigated the impact of vision encoder resolution on its proposed CCEval benchmark.
Among the three studied vision encoders (CLIP-ViT-L-112, CLIP-ViT-L-224, CLIP-ViT-L-336), higher resolution generally results in lower degrees of hallucination.
These works indicate that scaling up vision resolution is a straightforward yet effective solution.

\subsubsection{Versatile Vision Encoders}
Several studies~\cite{tong2024eyes,visual_experts,jain2023vcoder} have investigated vision encoders for MLLMs.
Typically, the CLIP ViT image encoder is used as the vision encoder in most MLLMs thanks to its remarkable ability to extract semantic-rich features.
However, CLIP has been shown to lose some visual details compared to pure vision models like DINO ViT~\cite{dino_vit}.
Therefore, recent studies have proposed complementing this information loss by incorporating visual features from other vision encoders.
The work of \cite{tong2024eyes} proposes mixing features from CLIP ViT and DINO ViT.
Specifically, it experimented with additive and interleaved features.
Both settings show that there is a trade-off between the two types of features.
A more dedicated mechanism is needed.

Concurrently, a visual expert-based model proposed in \cite{visual_experts} aims to mitigate the information loss caused by the CLIP image encoder.
Instead of merely mixing features, this paper enhances the visual perception ability of MLLMs by focusing on knowledge enhancement, relying on two pivotal modules: multi-task encoders and the structural knowledge enhancement module.
The multi-task encoders are dedicated to integrating various types of latent visual information extracted by multiple visual encoders.
Additionally, the structural knowledge enhancement module is designed to utilize visual tools, such as OCR tools and object detectors, to extract prior knowledge from visual inputs.

Following the approach of the structural knowledge enhancement module in~\cite{visual_experts}, another line of research investigates the utilization of vision tool models to enhance the perception of MLLMs.
VCoder~\cite{jain2023vcoder} utilizes additional perception formats, such as segmentation masks and depth maps, to enhance the object identification ability of the MLLM.
Another work~\cite{mitigate_detect_ocr} ensembles additional object detection and optical-character recognition models into the MLLM architecture.
It also explores various ways to integrate this information, including training-free infusion, LoRA~\cite{hu2021lora} augmented retraining, and LoRA augmented finetuning.
EAGLE~\cite{villa2025eagle} retrains the vision encoder with enhanced visual grounding capability, reducing hallucination in downstream application on MLLMs.

\subsubsection{Dedicated Module}
Following our previous discussion, the parametric knowledge embedded in the LLM is identified as a significant factor leading to hallucination, directing the generation to be based on language knowledge instead of visual content.
To address this issue, the work of \cite{halle_switch} proposes training a dedicated "\textit{switch}" module, termed \textit{HallE-Switch}, which controls the extent of parametric knowledge within detailed captions.
The detailed implementation is inspired by LM-switch~\cite{lm_switch}, which involves adding a control parameter $\epsilon$ serving as a "switching value".
The switch module is trained using contrastive training data from both contextual (visual content-related) and parametric datasets.
During inference, addressing hallucination can be attempted by tuning the control parameter $\epsilon$.

In the work of PATCH~\cite{PATCH}, it analyze the hallucination problem from an architectural perspective.
Correspondingly, it proposes a tuning strategy aligning visual and textual features in the semantic space to mitigate object hallucinations.
Concretely, it inserts several trainable and pluggable virtual tokens between image features and enhanced prompt texts, bridging the gap between the encoded image features and input augmented texts. During inference, the fine-tuned virtual token embeddings are added to the original vocabulary, making it a plug-and-play method.

\subsection{Training}
\label{sec:mitigating_training}

\subsubsection{Auxiliary supervision}
The primary supervision signal of training MLLMs is language modeling loss (implemented as \textit{CrossEntropyLoss}) in both pre-training and finetuning stage.
However, such supervision may not be sufficient to process the rich information encoded in the visual content.

\paragraph{\textbf{Visual Supervision.}}
Accordingly, the work of~\cite{RAH_bench} constructs a fine-grained vision instruction dataset based on Panoptic Scene Graph (PSG), called Relation-Associated Instruction (RAI-30k).
In addition to standard dialogues, each instruction in RAI-30k is associated with a relation annotation in PSG, which includes mask annotations for related instances.
With these additional annotations, it further supervises MLLMs with mask prediction loss using a state-of-the-art expert vision model, SAM~\cite{SAM}, guiding MLLMs to focus on highly-related image content.
With the additional supervision from the mask prediction loss, MLLMs are encouraged to extract features that can better represent these crucial instances, thus generating more accurate responses and mitigating vision hallucination.
The intuitive idea of supervising MLLMs with grounding shows promising performance in mitigating hallucination.

\paragraph{\textbf{Contrastive Learning.}}
Another line of work resort to contrastive learning to encourage better alignment at the embedding space~\cite{HACL_contrastive} or surpass hallucinatory response~\cite{halva} during training.

As introduced earlier, popular MLLMs typically project the encoded vision features into the input space of a specific LLM.
HACL~\cite{HACL_contrastive} argues that an ideal projection should blend the distribution of visual and textual embeddings.
However, despite visual projection, a significant modality gap exists between textual and visual tokens, suggesting that the current learned interfaces are not effective in mapping visual representations into the textual representation space of LLMs.
This issue potentially exacerbates the tendency for MLLMs to generate more hallucinations.
Therefore, HACL proposes enhancing the alignment between visual and textual representations through contrastive loss.
Texts with hallucinations are used as hard negative examples for image anchors.
The loss pulls representations of non-hallucinating text and visual samples closer while pushing representations of non-hallucinating and hallucinative text apart.

HALVA~\cite{halva} utilizes the philosophy of contrastive learning in the language generation process.
Specifically, it first uses generative data augmentation to construct a training set of `hallucinated' and `correct' response pairs, by selectively altering the ground-truth phrases in the correct responses, while keeping the overall structure intact.
Then, a phrase-level alignment loss is proposed to finetune the MLLM on the constructed correct and hallucinated response pairs.
The data and loss provides a fine-grained supervision on phrase-level, effectively reducing hallucination.

VISTA~\cite{VISTA} address the problem from the perspective of visual information loss in MLLMs.
It introduces two complementary modules: Visual Steering Vector (VSV) that counteracts gradual visual information loss by extracting and reinforcing visual cues in activation space, and Self-Logits Augmentation (SLA) which utilizes early excitation patterns to prioritize semantically meaningful tokens.
In implementation, it involves a positive prompt and a negative prompt (without image tokens), forming a contrastive-style learning paradigm.

\paragraph{\textbf{Others.}}
Recalling the work of EOS Decision~\cite{eos_token}, to teach the model to terminate the generation process properly, this work also designs a learning objective, termed Selective EOS Supervision, in addition to the data filtering strategy.
This is achieved by simply modifying the Maximum Likelihood Estimation (MLE), enabling the model to mitigate hallucination through learning from regular instruction data.

Recently, the work of CCA~\cite{CCA} discusses a novel perspective regarding positional dependency modeling.
The authors observe that due to the long-term decay in RoPE, LVLMs tend to hallucinate more when relevant visual cues are distant from instruction tokens in the multimodal input sequence.
Therefore, they propose Concentric Causal Attention (CCA), a positional alignment strategy that mitigates the impact of RoPE long-term decay in LVLMs by naturally reducing relative distance between visual and instruction tokens.

\subsubsection{Reinforcement Learning}

Reinforcement learning (RL) is introduced to train MLLMs for mitigating hallucinations by conducting the following perspectives: 1) Automatic Metric-based Optimization, 2) Reinforcement Learning from AI Feedback, 3) Reinforcement Learning from Human Feedback.

\paragraph{\textbf{Automatic Metric-based Optimization}}
Motivated by the limitation of LLMs (and MLLMs) training, which is unable to optimize at the sequence level, the MOCHa~\cite{ben2023mocha_openchair} framework is proposed to apply reinforcement learning.
This work aims to improve the accuracy and relevance of image captioning, thereby reducing hallucination.
The framework introduces three metric-based objectives to guide the reinforcement learning process for image captioning:
1) Natural Language Inference (NLI) for fidelity, focusing on the accuracy of the caption in describing the image content;
2) BERTScore~\cite{zhang2019bertscore} for semantic adequacy, assessing the relevance and richness of the description;
and 3) Kullback–Leibler (KL) divergence for regularization, which constrains the model to stay close to its initial policy.
The framework incorporates these objectives into a multi-objective reward function for reinforcement learning.
Subsequently, the proximal policy optimization reinforcement learning algorithm is employed to maximize the expected reward.
By promoting the creation of accurate, contextually appropriate, and varied descriptions, the hallucination of MLLM can be mitigated.

\paragraph{\textbf{Reinforcement Learning from AI Feedback (RLAIF)}}
HA-DPO~\cite{HA_DPO} addresses hallucination as a preference selection problem by training models to prioritize accurate responses over hallucinatory ones.
To achieve this goal, HA-DPO initially constructs a high-quality dataset.
Specifically, it first utilizes MLLMs to generate descriptions corresponding to images, then employs GPT-4 to detect whether these descriptions contain hallucinations.
If hallucinations are detected, the descriptions are rewritten.
Thus, HA-DPO constructs a dataset that includes both accurate descriptions (positive samples) and hallucinatory descriptions (negative samples).
HA-DPO then trains the model using these sample pairs, enabling it to distinguish between accurate and hallucinatory descriptions.
This goal is achieved through direction preference optimization (DPO), which optimizes a specific loss function designed to maximize the model's preference for positive samples while minimizing its preference for negative samples.

A concurrent work, Silkie~\cite{dpo_silkie}, introduces a similar approach of utilizing preference-based reinforcement learning to enhance the faithfulness of MLLMs.
Specifically, it emphasizes the concept of reinforcement learning from AI feedback (RLAIF) by distilling preferences from a more robust MLLM, \textit{i.e.}, GPT-4V~\cite{gpt4v_card}.
Responses are first generated by models from 12 MLLMs, and then assessed by GPT-4V.
The constructed dataset, termed as VLFeedback, contains preferences distilled from GPT-4V and is utilized to train other MLLMs through direct preference optimization.

POVID~\cite{dpo_povid}, challenges the assumption underlying previous DPO-based methods.
These methods rely on the traditional preference data generation process in LLMs, where both preferred and dispreferred responses may potentially be incorrect.
Therefore, this work proposes the Preference Optimization in VLLM with AI-Generated Dispreferences (POVID) framework, aiming to exclusively generate dispreferred feedback data using AI models.
The dispreferred data is generated by: 1) utilizing GPT-4V to introduce plausible hallucinations into the answer, and 2) provoking inherent hallucination by introducing noise into MLLMs.
In the DPO optimization framework, the ground-truth multimodal instructions serves as the preferred answers.

RLAIF-V~\cite{yu2024rlaif_v} argues that the most existing RLAIF frameworks rely on expensive proprietary models, limiting the scalability. To bridge the gap, this work proposes a solution to utilize fully open-sourced MLLM to generate high-quality feedback.
CLIP-DPO~\cite{clip_dpo} takes one step further by getting rid of offline data collection.
Instead, starting from the initial pool of supervised fine-tuning data, it generates a diverse set of predictions, which are ranked based on their CLIP image-text similarities, and then filtered using a robust rule-based approach to obtain a set of positive and negative pairs for DPO-based training.

Following the path of RLAIF, subsequent works, including FGAIF~\cite{jing2024fgaif} and HSA-DPO~\cite{hsa_dpo} further enhance the model by proposing fine-grained reward (feedback), including multiple types (object, attribute, relation) and segment-level reward, etc.
V-DPO~\cite{v_dpo} devises a vision-guided direct preference optimization with a synthetic dataset containing both response-contrast and image-contrast preference pairs.
HDPO~\cite{hdpo} develop three types of preference pair data targeting the following causes of MLLM hallucinations: 1) insufficient visual capabilities; 2) long context generation, and 3) multimodal conflicts.
OPA-DPO~\cite{opa_dpo} identifies a crucial factor of DPO: outcomes are largely contingent on whether the constructed data aligns on-policy w.r.t the initial (reference) policy of DPO.
To alleviate the problems, it proposes OnPolicy Alignment (OPA)-DPO framework, which uniquely leverages expert feedback to correct hallucinated responses and aligns both the original and expert-revised responses in an on-policy manner.

\paragraph{\textbf{Reinforcement Learning from Human Feedback (RLHF)}}

HalDetect~\cite{HalDectect_gunjal2023detecting} first introduces the M-HalDetect dataset for detecting hallucinations, which covers a wide range of hallucinatory content, including non-existent objects, unfaithful descriptions, and inaccurate relationships.
It then proposes a multimodal reward model to detect hallucinations generated by MLLMs.
The reward model is trained on the M-HalDetect dataset to identify hallucinations in the generated text.
To utilize the trained reward model to reduce hallucinations, the authors introduced Fine-grained Direct Preference Optimization (FDPO).
FDPO uses fine-grained preferences from individual examples to directly reduce hallucinations in generated text by enhancing the model's ability to distinguish between accurate and inaccurate descriptions.

LLaVA-RLHF \cite{llava_rlhf_sun2023aligning} also try to involve human feedback to mitigate hallucination.
It extends the RLHF paradigm from the text domain to the task of vision-language alignment, where human annotators were asked to compare two responses and pinpoint the hallucinated one.
The MLLM is trained to maximize the human reward simulated by an reward model.
To address the potential issue of \textit{reward hacking}, \textit{i.e.,} achieving high scores from the reward model does not necessarily lead to improvement in human judgements, it proposes an algorithm named Factually Augmented RLHF.
This algorithm calibrates the reward signals by augmenting them with additional information such as image captions.

Similarly, RLHF-V~\cite{yu2023rlhf_v} also employs the RLHF paradigm to enhance the pre-trained MLLM.
Specifically, this work emphasizes two improvements:
1) at the data level, it proposes to collect human feedback in the form of fine-grained segment-level corrections, providing a clear, dense, and fine-grained human preference.
2) at the method level, it proposes dense direct preference optimization (DDPO) that directly optimizes the policy model against dense and fine-grained segment-level preference.

Another similar work, ViGoR~\cite{reward_model_yan2024vigor}, also designs a fine-grained reward model to update pre-trained MLLMs, aiming to improve visual grounding and reduce hallucination.
The reward modeling in this work encompasses both human preferences and automatic metrics.
Specifically, it collects human judgment and preferences for the responses generated by MLLMs by asking crowd-workers to provide fine-grained feedback at the sentence level.
The collected human preference data is used to train a reward model.
Additionally, it leverages advanced vision perception models to automatically score the grounding and fidelity of the text generated by an MLLM.
Both sources are combined into a single reward score during the reinforcement learning procedure.

\paragraph{\textbf{Visual Generative Feedback.}}
The advancement of text-to-image (T2I) generation models~\cite{stable_diffusion} offers new opportunities in mitigating hallucination.
ESREAL~\cite{ESREAL} proposes to utilize a T2I model to semantically reconstruct an image from the generated caption.
After that, the semantic misalignment between the two images can serve as feedback to optimize the model.
In the work of ESREAL~\cite{ESREAL}, the optimization is achieved through proximal policy
optimization~\cite{ppo}.
Such visual generative feedback has also been utilized during inference time to provide visual contrast signal for decoding, as explored in ConVis~\cite{park2024convis}.

Another relevant work~\cite{DeGF} introduces self-correcting Decoding with Generative Feedback (DeGF), a training-free algorithm that incorporates feedback from text-to-image generative models into the decoding process to effectively mitigate hallucinations in MLLMs.
Specifically, DeGF generates an image from the initial response produced by MLLMs, which acts as an auxiliary visual reference and provides self-feedback to verify and correct the initial response through complementary or contrastive decoding.
This is built based on the assumption: given a visual input and a textual prompt to an MLLM, if the generated response conditioned on the original image is accurate and non-hallucinatory, a text-to-image generative model should be capable of reversing this process to produce a similar image from that response.

\subsubsection{Unlearning}
Unlearning refers to a technique designed to induce a model to 'forget' specific behaviors or data, primarily through the application of gradient ascent methods~\cite{unlearning}.
Recently, unlearning for LLMs has been receiving increasing attention~\cite{unlearn_llm_privacy}, effectively eliminating privacy vulnerabilities in LLMs.
In the context of MLLMs, a recent work~\cite{xing2024efuf} introduces the Efficient Fine-grained Unlearning Framework (EFUF), applying an unlearning framework to address the hallucination problem.
Specifically, it utilizes the CLIP model to construct a dataset comprised of both positive samples and negative (hallucinated) samples.
The training loss is applied separately for positive and negative at the sub-sentence level.
To the best of our knowledge, EFUF~\cite{xing2024efuf} is the first and only work that applies the unlearning framework to the task of hallucination mitigation, opening up a new path for future research.

\subsection{Inference}
\label{sec:mitigating_inference}

\subsubsection{Generation Intervention}

\paragraph{\textbf{Contrastive Decoding}}
Contrastive decoding~\cite{li2022contrastive,chuang2023dola} has been a popular technique in large language models to reduce hallucination and promote the plausibility of the generated text.
In the language domain, the differences between a large "expert" language model (LM) and a smaller "amateur" LM are leveraged.
The core idea is to identify and prefer outputs where the expert model's confidence significantly exceeds that of the amateur model.

In the multimodal domain, instead of using "amateur" models, there are efforts on building a normal distribution and a "distorted" distribution, and then conduct contrastive decoding.
One representative work is VCD (Visual Contrastive Decoding)~\cite{VCD}, which is designed to suppress the statistical biases and language priors in MLLMs during the decoding phase.
The main assumption of VCD is that a distorted visual input would lead to text responses with more biases and priors.
Thus, by contrasting output distributions derived from original and distorted visual inputs, VCD aims to effectively reduce the over-reliance on statistical bias and language priors.
Specifically, the decoding probability distribution is calibrated using the reference (distorted) distribution.

There are many subsequent works on improving contrastive decoding to reduce hallucination.
Thus, we do not dive into the details of each specific work.
In the work of ICD~\cite{ICD}, the distorted distribution is achieved by manipulating the text instruction prompt.
Specifically, role prefixes are appended to the instructions to form disturbance instructions, which can exacerbate hallucinations, as observed by the authors.
With the two distributions, contrastive decoding can be performed to reduce hallucination.
HIO~\cite{HIO} proposes a Hallucination-Induced Optimization strategy that seeks to amplify the contrast between hallucinatory and output tokens relying on an `Evil-MLLM' intentionally producing hallucination.
CODE~\cite{kim2024code} utilizes the comprehensive descriptions from model itself as visual counterpart to build contrastive decoding, aiming to correct and improve response alignment with actual visual content.
VACoDe~\cite{kim2024vacode} explores how to utilize multiple image augmentations in contrastive decoding and proposes to adaptively selects the augmentation with the highest contrast for each task using the proposed softmax distance metric.
SID~\cite{SID} aims to induce hallucination by proposing token-level disturbances.
This strategy preserves only the least important vision tokens after the early decoder layers, thereby adaptively amplify vision-and-text association hallucinations during auto-regressive decoding.
This strategy ensures that multimodal knowledge absorbed in the early decoder layers induces multimodal contextual rather than aimless hallucinations.
DAMRO~\cite{gong2024damro} takes inspiration from the analysis of the attention pattern in the vision encoder and MLLM, and finds that the distributions tend to focus on particular background tokens rather than the referred objects in the image, potentially introducing hallucination.
To address the issue, it proposes to employ classification token (CLS) of ViT to filter out high-attention outlier tokens scattered in the background and then eliminate their influence by contrastive decoding.
VaLiD~\cite{wang2024valid} utilizes the technique of contrastive decoding from a vision encoder perspective.
It first identifies that hallucinations in MLLMs may originate at the initial stage of visual perception in vision encoder.
Based on this, it utilizes information from the distorted visual layers and applies contrastive decoding.
CMVED~\cite{CMVED}, when build contrastive distribution, performs distortion by selectively masking the value vectors associated with high cross-modal attention weights in self-attention layers, which suppresses important inter-modality correlations while retaining the spurious ones in distorted distribution.
VASparse~\cite{zhuang2025vasparse} levearges the observation of sparse activation of attention in MLLMs and designs a token sparsification strategy that balances efficiency and trustworthiness.
Based on that, it further introduces a sparse-based visual contrastive decoding method to recalibrate the distribution of hallucinated outputs.
Octopus~\cite{suo2025octopus} argues that the mechanism of hallucination occurrence is a complex hybrid and different samples (or tokens), while current contrastive decoding methods mostly apply the same disturbed manner for all samples and generative steps.
Thus, it focuses on guiding the model to dynamically organize the contrastive decoding workflow and selecting the appropriate contrastive decoding strategy based on different inputs.

\paragraph{\textbf{Guided Decoding}}

MARINE~\cite{cfg_mitigate} proposes a training-free approach.
It employs an additional vision encoder for object grounding and utilizes the grounded objects to guide the decoding process.
Specifically, it innovatively adapts the classifier-free guidance~\cite{cfg} technique to implement guided decoding, showing promising performance in emphasizing the detected objects while reducing hallucination in the text response.

Similarly, GCD~\cite{deng2024seeing} devises a CLIP-Guided Decoding (GCD) approach.
It first verifies that CLIPScore~\cite{CLIP} can effectively distinguish between hallucinated and non-hallucinated sentences through a series of studies across different models and datasets.
Based on this conclusion, it further recalibrates the decoding process of MLLMs, including two steps: 1) reliability scoring, which designs a (CLIP-based) scoring function aiming to assign higher scores to candidate responses that are less likely to be hallucinated, and 2) guided sentence generation, which generates responses based on this scoring.
This is implemented in a similar way to beam search but at the sentence level.

HALC~\cite{chen2024halc} provides a key insight that when decoding a specific token in the MLLM, identifying a token-wise optimal \textit{visual context} to provide the most informative visual grounding can effectively reduce hallucination.
\textit{Visual context} refers to the visual tokens that can be grounded from the generated text response.
An oracle study showed that decoding from the provided optimal visual contexts eliminates over 84.5\% of hallucinations.
Based on the insight and observation, the authors designed mechanisms to locate the fine-grained visual information to correct each generated token that might be hallucinating.
This is essentially a visual content-guided decoding strategy.
In addition to token-level correction, HALC also incorporates a \textit{matching-based beam search} that utilizes a visual matching score to steer the generation of the final outputs, balancing both object hallucination mitigation and text generation quality.

The work of DeCo~\cite{deco} has conducted an empirical analysis and find that MLLMs are not blind; they can recognize objects in the preceding layers, but this recognition is suppressed in later layers, leading to hallucinations.
Based on this observation, it proposes Dynamic Correction Decoding with preCeding-Layer Knowledge (DeCo) to mitigate hallucinations for MLLMs.
It dynamically selects preceding layer and utilizes its prior knowledge as a guidance to correct the final output logits.

SumGD~\cite{sum_gd} aims to address hallucination caused by language prior.
The approach employs a summarization technique that selectively retains essential information from previously generated sentences, encouraging LVLMs to more effectively incorporate image information.
To minimize unnecessary intervention for preserving text quality, the summarization is referenced only when predicting image-related POS tokens, which require image-specific details.

\paragraph{\textbf{Visual Amplification}}
Many works \cite{M3ID} have identified that one cause of hallucination is the weak dependency on the visual prompt during generation.
And this dependency even decreases as more tokens are generated~\cite{M3ID}, which is consistent with the observation of \textit{longer output tend to generate more hallucination} in the work of \cite{han2024skip}.
To address this, M3ID~\cite{M3ID} proposes a novel decoding algorithm, which models two output distributions with and without the image prompt as condition.
During generation, the final output is determined together by two distributions, explicitly amplifying the image prompt.

IBD~\cite{zhu2024ibd} proposes an image-biased decoding strategy, which is also closely related to contrastive decoding.
Specifically, IBD involves computing a more reliable next-token probability distribution by contrasting the predictions of the original model with those of an image-biased model, which focuses more on the image information.
The image-based model is created by modifying the attention weight matrix structure within the original model, without altering its parameters.
This approach emphasizes the knowledge of the image-biased model and diminishes that of the original model, which may be text-biased.
Thus, it encourages the extraction of correct content while suppressing hallucinations resulting from textual over-reliance.

AGLA~\cite{an2024agla} argues that current MLLMs demonstrate attention deficiency that causes the ignorance of prompt-relevant object features in images.
Therefore, it introduces prompt-relevant local attention to generate an augmented view of the image for hallucination mitigation.
Specifically, it design an image-prompt matching scheme that computes the prompt-relevant attention and further generates an augmented view of the original image, better supporting object-related queries.

RVD~\cite{mmhal_snowball}(Residual Visual Decoding) proposes a straightforward strategy.
By residual connecting the visual information and the current user instruction, distributions that emphasizing the visual information are derived to revise the original output distribution.
This strategy has shown effectiveness on amplifying the visual information.

Another representative work PAI~\cite{PAI} (Pay Attention to Image) intervenes in the inference process to make it more image-centric, following the original image perception direction.
To achieve this, the method focuses on the self-attention heads in the decoder layers of MLLMs.
It enhances the attention weights for image tokens in their original directions during inference.
This allows the model to use the updated attention matrix to calculate the hidden states for the generated token, thereby incorporating more consideration for image representation during the generation process.

EAH~\cite{EAH} also identifies this problem of visual attention sink.
To address this, it proposes a training-free method named Enhancing Attention Heads (EAH), an approach designed to enhance the convergence of image tokens attention sinks in the shallow layers.
EAH identifies the attention head that shows the vision sink in a shallow layer and extracts its attention matrix.
This attention map is then broadcast to other heads in the layer, thereby strengthening the layer to pay more attention to the image itself.
A similar idea has also been explored in VHR~\cite{VHR}, a training-free approach aimed at enhancing the model’s reliance on visual context rather than language priors.
This achieved by first identifying key attention heads based on their VHD scores and then amplifying their contributions during generation.

PM~\cite{magnifying} proposes an intuitive strategy that adaptively modulates spatial resolution, processing key visual regions at higher detail while compressing less relevant areas.
It refines the visual input by magnifying relevant regions while compressing less critical areas, implemented as a resampling process.

\paragraph{\textbf{Others}}
The work of OPEAR~\cite{huang2023opera} makes an interesting observation that most hallucinations are closely tied to the knowledge aggregation patterns manifested in the self-attention matrix, \textit{i.e.,} MLLMs tend to generate new tokens by focusing on a few summary tokens rather than all the previous tokens.
Such a partial over-trust inclination results in neglecting image tokens and describing the image content with hallucination.
Based on this observation, a decoding method for MLLMs grounded in an \textbf{O}ver-trust \textbf{P}enalty and a \textbf{R}etrospection-\textbf{A}llocation strategy is proposed.
First, a penalty term on the model logits is introduced during the MLLM beam-search decoding process to mitigate the over-trust issue.
Additionally, to handle the hard cases that cannot be addressed by the penalty term, a more aggressive strategy called the rollback strategy is proposed to retrospect the presence of summary tokens in the previously generated tokens and reallocate the token selection if necessary.
A follow-up work HELPD~\cite{yuan2024helpd} modified the penalty by incorporating visual attention into the penalty score computation and makes the final logits place more emphasis on the image input, amplifying the visual attention as discussed above.

Another interesting study observes that the hallucination of MLLMs seems to be easily triggered by paragraph break `\textbackslash{}n\textbackslash{}n' \cite{han2024skip}.
Based on this observation, this work proposes two simple methods to reduce hallucination by avoiding generating `\textbackslash{}n' during generation.
First, intuitively, users can design the prompt to instruct the model to output responses within one paragraph, avoiding `\textbackslash{}n'.
Besides, the authors tried to alter the output logits during generation by manually lowering the probability of generating `\textbackslash{}n'.
Experimental results show that this simple strategy can alleviate hallucination on popular benchmarks.

The work of ProjectAway~\cite{vl_interp} explores a novel strategy that directly edits the latent image representation.
Through analysis and experiment, they find the explored strategy can erase both real and hallucinated objects with high rates of removal.
Therefore, this tool is utilized to detect hallucinations and reduce hallucination by erasing hallucinations from the VLM’s internal representations.
Similar to editing latent space, Nullu~\cite{yang2024nullu} directly edits the model weights, based on an unsafe subspace, which is called HalluSpace in the paper.
By orthogonalizing the model weights, input features will be projected into the Null space of the HalluSpace to reduce hallucination.

CausalMM~\cite{causal_mm} develops a causal inference framework that applies structural causal modeling to MLLMs, treating modality priors as a confounder between attention mechanisms and output.
By applying counterfactual reasoning on the framework, it makes the output more aligned with multimodal inputs.

Some works propose various prompting strategies to mitigate hallucination.
The work of ~\cite{kim2024if} proposes to instruct the MLLM first think about some counterfactual contents. Then, at the actual generation step, the model should answer the user query while simultaneously avoid the counterfactual terms.
VIC~\cite{VIC} designs a thinking before looking strategy to leverage the reasoning capability of the LLM, enabling MLLMs to anticipate and recognize patterns more efficiently by dynamically adjusting reasoning steps.

\subsubsection{Visual Prompting}
As many works~\cite{huang2023opera,M3ID} have identified that one prominent cause of hallucination is weak or fading visual attention, one intuitive way to alleviate this issue is to explicitly highlight the objects-of-interests.
An early exploration, SoM-LLaVA~\cite{som_llava}, applies the Set-of-Mask prompting strategy~\cite{som_prompting} on MLLMs, demonstrating effectiveness on reducing hallucination.

The work of VAP~\cite{VAP} designs an solution: strategically crafted perturbation to visual inputs can redirect LVMs’ decision-making processes away from parametric biases without altering the original model’s architecture or mechanisms.
It formulates an adversarial strategy to align the MLLM responses with visual content while reducing the impact of parametric knowledge bias.
Eventually, the perturbation noise is optimized to alleviate parametric bias thus reducing hallucination.
Essentially, this strategy can be regarded as a type of visual prompting.

Another work~\cite{tomita2025role} investigates the role of background information in reducing object hallucination from a visual prompting perspective.

\subsubsection{RAG}
Recently, in the realm of LLM, Retrieval-Augmented Generation (RAG)~\cite{wang2020show,rag_nlp} from external knowledge resources has shown promise in reducing language hallucinations.
This retrieval augmentation serves as a flexible way to extend beyond the model’s inherent knowledge without the burden of extensive training costs.
This technique has also been used to mitigate hallucination in MLLM.
ARA~\cite{ara} adopts RAG by 1) decomposing the target object causing hallucination from the input image; 2) discerning the most effective retrieval technique and securing trustworthy results, considering the nature of multimodal; 3) initiating retrieval only during periods of low LVLM certainty and knowledge deficiency to reduce cost.
A similar idea is further explored in a more recent work FilterRAG~\cite{RAG}.

\subsubsection{Ensembling}
As a black box model, MLLM can be affected by even smaller perturbation in at the input model parameters, which may lead to hallucination.
Therefore, it is intuitive to consider ensembling results from multiple inputs or models to obtain a more robust result.

RITUAL~\cite{woo2024ritual} proposes to leverage random image transformations to complement the original image and enhance models’ robustness.
The key idea is that by exposing the model to diverse visual transformations—such as changes in orientation, scale, and color—during decoding, it can better discern the true contents of the original image and reduce the likelihood of generating hallucinatory outputs.

MAD~\cite{multi_agent_debate} addresses the hallucination problem from a model ensembling-style solution.
Specifically, it adopts a multi-agent debate strategy by instantiating several MLLM as agents to conduct a multi-round debate.
This cross-agent debate encourages divergent-thinking and promotes cognitive synergy.

Another ensembling-style solution is explored in MVP~\cite{MVP}, where a Multi-view-Multi-path strategy is proposed.
It involves a multi-view information-seeking strategy that exhausts the perception of the image from varying dimensions: ``top-down'', ``bottom-up'', and ``regular''.
In addition, during the answer decoding stage, it further introduces multi-path reasoning for each information view by explicitly quantifying the certainty score of
the potential answers and then aggregating the overall certainty among multiple paths.

The work of Dentist~\cite{dentist} argues that there are different types of queries that can be addressed by different strategies.
Therefore, it proposes a unified framework that first classifies the queries, then performs different processes of hallucination mitigation based on the classification result.

\subsubsection{Post-hoc Correction}

Post-hoc correction refers to first allowing the MLLM to generate a text response and then identifying and eliminating hallucinating content, resulting in less hallucinated output.
This is usually achieved by grounding on visual content~\cite{yin2023woodpecker}, pre-trained revisior~\cite{LURE_zhou2023analyzing}, and self-revision~\cite{lee2023volcano}.

Woodpecker~\cite{yin2023woodpecker} is an early attempt on hallucination detection and correction.
Similar to how a woodpecker heals trees, Woodpecker picks out and corrects hallucinations from the generated text.
The key idea of Woodpecker is to extract key concepts from the generated text and validate them using visual content.
Subsequently, the hallucinated concepts can be detected and corrected accordingly.
Specifically, it consists of five stages:
1) \textit{Key concept extraction} identifies the main objects mentioned in the generated sentences;
2) \textit{Question formulation} asks questions around the extracted objects;
3) \textit{Visual knowledge validation} answers the formulated questions via expert models;
4) \textit{Visual claim generation} converts the above Question-Answer (QA) pairs into a visual knowledge base;
5) \textit{Hallucination correction} modifies the hallucinations and adds the corresponding evidence
under the guidance of the visual knowledge base.
Woodpecker is a training-free method, where each component can be implemented using either hand-crafted rules or off-the-shelf pre-trained models.

Another line of work rectifies the generated text using a dedicatedly trained revisor model.
Specifically, inspired by denoising autoencoders~\cite{denoising_autoenc}, which are designed to reconstruct clean data from corrupted input, LURE~\cite{LURE_zhou2023analyzing} employs a hallucination revisor that aims to transform potentially hallucinatory descriptions into accurate ones.
To train such a revisor model, a dataset has been constructed.
Each example in this dataset consists of an image accompanied by a hallucinatory description, with the correct description serving as the target output.
The hallucinatory descriptions are generated by modifying the accurate descriptions using GPT-3.5.
These adjustments are guided by factors related to object hallucination, including co-occurrence, object uncertainty, and object position.
After that, the authors fine-tuned an MLLM using this dataset to serve as a revisor, which is used as an additional step for rectifying the output of an MLLM during generation.

Similar idea has also been explored in Volcano~\cite{lee2023volcano}.
It introduces a self-revising mechanism to reduce hallucination.
It consists of four stages: 1) generate initial response; 2) generate feedback for the initial response; 3) revise the response using this feedback; 4) compare the responses before and after revision to decide which one is better.
Stages 2-4 are repeated iteratively.
To provide better feedback and decision-making, the model is fine-tuned on a curated dataset.
The dataset is organized using ChatGPT.

LogicCheckGPT~\cite{logic_check} is a self-revising-based hallucination mitigation method.
Unlike Volcano~\cite{lee2023volcano}, which revises the generated response with the help of \textit{general} feedback, LogicCheckGPT delves into the \textit{logical consistency} of MLLMs' responses.
Specifically, the approach can be formulated into two stages: the first stage involves inquiring attributes of objects, followed by inquiring objects based on attributes.
Whether their responses can form a logical closed loop serves an indicator of object hallucination.
If the ratio of closed loops to the total number of questions exceeds a certain threshold, rectify the hallucinated objects by prompting the MLLM.

VFC~\cite{VFC} adopts an agent-style workflow to generate accurate, detailed caption.
This is achieved by 1) proposal, where image-to-text captioning models propose multiple initial captions;
2) verification, where a large language model (LLM) utilizes tools such as object detection and VQA models to fact-check proposed captions;
3) captioning, where an LLM generates the final caption by summarizing caption proposals and the fact check verification results.
The verification step can effectively mitigate hallucination.
Dhall~\cite{wu2024combating} designs a holistic reasoning framework to mitigation hallucination, encompassing six reasoning modules.

\section{Challenges and Future Directions}
\label{sec:future}

The research of hallucination in MLLMs is still at early stage, remaining a variety of research problems to be explored.
In this section, we delve into the challenges and future directions of this pivotal domain.

\subsection{Data-centric Challenges and Innovations}
The reliance of MLLMs on large volumes of data presents significant challenges in terms of data quality, diversity, and bias.
In Sec.~\ref{sec:causes_data}, previous works have identified several core issues that may cause hallucination.
In order to improve the accuracy and reliability of hallucinated content, it is crucial to ensure that MLLMs have access to high-quality and diverse training data.
Future research should focus on developing techniques for data collection, augmentation, and calibration.
Firstly, collecting enough data at the initial stage is crucial to address the data scarcity issue and increase data diversity.
Secondly, data augmentation is an effective solution to further expand the size of data.
Finally, exploring methods for re-calibrating existing datasets is crucial.
This includes eliminating biases, promoting diversity and inclusivity, and mitigating other potential issues that may induce hallucinations.

\subsection{Cross-modal Alignment and Consistency}
The key challenge of multimodal hallucination is the cross-modal consistency issue.
Ensuring that generated content remains consistent and contextually relevant to the input modality requires sophisticated techniques for capturing and modeling cross-modal relationships.
The direction of cross-modal alignment encompasses both MLLMs training and hallucination evaluation.
Regarding training, future research should explore methods for aligning representations between different modalities.
Achieving this goal may involve designing more advanced architectures, introducing additional learning objectives~\cite{HACL_contrastive}, or incorporating diverse supervision signals~\cite{RAH_bench}.
Regarding evaluation, cross-modal consistency checking has been a long-standing topic, ranging from multimodal understanding~\cite{CLIP,blip-2} to text-to-image generation~\cite{dsg_t2i, easy_detect}.
Drawing on proven experiences from these domains to improve the assessment of MLLM hallucination, or unifying them into an overall framework, may be promising research directions.

\subsection{Advancements in Model Architecture}
Despite recent advancements in model architectures of LLMs and MLLMs, designing effective architectures specifically tailored to hallucination remains a challenge.
Developing advanced model architectures capable of capturing complex linguistic structures and generating coherent and contextually relevant output based on input visual content is essential for improving the performance of MLLMs.
Future research can explore innovative architectural designs based on identified causes of hallucination. This includes developing stronger visual perception models, innovative cross-modal interaction modules capable of transferring cross-modal information seamlessly, and novel large language model architectures faithful to input visual content and text instructions, etc.

\subsection{Establishing Standardized Benchmarks}
The lack of standardized benchmarks and evaluation metrics poses significant challenges in assessing the degree of hallucination in MLLMs.
In Table~\ref{tab:summary_bench_metrics}, it can be observed that there is a variety of evaluation benchmarks, but a lack of unified standards.
Among them, one of the most popular benchmarks might be POPE~\cite{pope}, which employs a 'Yes-or-No' evaluation protocol.
However, this binary-QA manner does not align with how humans use MLLMs.
Accordingly, some benchmarks specifically evaluate the hallucination of MLLMs in the (free-form) generative context.
Yet, they often rely on external models, such as vision expert models or other LLMs, which limits their widespread application.
Moving forward, future research can investigate standardized benchmarks that are theoretically sound and easy to use.
Otherwise, research on methods to mitigate hallucinations may be built on an incorrect foundation.

\subsection{Reframing Hallucination as a Feature}
Recently, discussions on social media~\cite{hallu_feature} have suggested that hallucination can be regarded as an inherent feature of LLMs and MLLMs.
The models are like dream machines.
Human users direct their dreams with prompts.
The prompts start the dream, and based on the model's hazy recollection of its training documents, most of the time the result goes someplace useful.
It's only when the dreams enter deemed factually incorrect territory that we label them as 'hallucinations'.
From this perspective, leveraging hallucination capabilities as a feature in downstream applications presents exciting opportunities for enhancing user experiences and enabling new use cases.
As humans are the end-users of these models, the primary goal is to enrich human user experiences.
Future research may switch the optimization objective from specific cross-modal benchmarks to human experience.
For example, Some content may cause hallucinations but will not affect the user experience, while some content may.
Alternatively, integrating hallucination to inspire more creative ideas in real-world applications could also be intriguing.

\subsection{Enhancing Interpretability and Trust}
Existing methods for hallucination mitigation are primarily based on empirical observations of specific patterns, such as skipping the `\textbackslash{}n' token and penalizing over-trust tokens.
However, despite the impressive improvements achieved on specific benchmarks, understanding the underlying mechanisms and decision-making processes remains challenging.
Future research should focus on developing techniques for interpreting and explaining the generation process of MLLMs, thereby providing insights into the factors influencing hallucinated content.
This includes investigating methods for visualizing model internals, identifying salient features and linguistic patterns, and tracing the generation process from input to output.
Enhancing the interpretability of MLLMs will not only improve our understanding of model behavior but also enable users to better assess hallucinated content in practical applications.

\subsection{Navigating the Ethical Landscape}
As MLLMs become increasingly proficient at generating realistic text, ethical considerations surrounding the use of generated content become paramount.
Especially in the context of hallucination, the generated response may contain severely concerning ethical content, amplifying the importance of the problem.
Addressing ethical concerns related to misinformation, bias, privacy, and societal impact is crucial for promoting responsible AI practices in the development and deployment of MLLMs.
In addition to addressing typical object hallucination, future research on MLLM hallucinations should prioritize ethical considerations throughout the entire lifecycle of MLLM development, from data collection and model training to deployment and evaluation.

\section{Conclusion}
\label{sec:conclusion}
Based on powerful large language models, multimodal large language models demonstrate remarkable performance across various multimodal tasks.
However, the phenomenon of hallucination presents a significant challenge to the practical applications of MLLMs, giving rise to undeniable concerns about safety, reliability, and trustworthiness.
In this comprehensive survey, we conducted a thorough examination of hallucinations within multimodal large language models, focusing on their underlying causes, evaluation metrics, benchmarks, and mitigation methods.
Despite considerable progress, hallucination remains a complex and persistent concern that warrants ongoing investigation.
The challenge of hallucination in multimodal large language models remains compelling, requiring continuous scrutiny and innovation.
In light of these challenges, we have outlined several promising future directions in this burgeoning domain.
Through navigating the intricate landscape of hallucinations, we aim for this survey to serve as a foundational resource for addressing the complexities of hallucination phenomena in MLLMs.
We envision this survey empowering researchers and practitioners to dedicate efforts to advancing research and developing robust solutions in this vital area of study.

\begin{acks}
This research is supported by the National Research Foundation, Singapore under its AI Singapore Programme (AISG Award No: AISG3-RP-2022-030).
\end{acks}

\bibliographystyle{ACM-Reference-Format}
\bibliography{llm,mllm}

\end{document}